%% file: Arxiv_main_IEEE.tex
\documentclass[journal]{IEEEtran}
\ifCLASSINFOpdf
\else
\fi
\usepackage{hyperref}
\usepackage{comment,color}
\usepackage{amsfonts}
\usepackage{amsmath}
\usepackage{multirow}
\usepackage{makecell}
\usepackage{subcaption}
\usepackage{mathtools}
\usepackage[ruled,vlined]{algorithm2e}
 \usepackage{relsize}
 \usepackage{graphics}
 \usepackage{wrapfig}

\DeclarePairedDelimiter{\ceil}{\lceil}{\rceil}
\usepackage{array}
\newcolumntype{H}{>{\setbox0=\hbox\bgroup}c<{\egroup}@{}}
\newcommand{\boldx}{\mbox{${\mathbf x}$}}
\newcommand{\bolda}{\mbox{${\mathbf a}$}}

\begin{document}
%
\title{Towards Interpretable-AI Policies Induction using Evolutionary Nonlinear Decision Trees for Discrete Action Systems}
%
%
%

\author {
        Yashesh Dhebar,
        Kalyanmoy Deb,
        Subramanya Nageshrao, Ling Zhu, and Dimitar Filev
\thanks{Yashesh Dhebar and Kalyanmoy Deb are with Michigan State University, East Lansing, MI, 48824 USA}
\thanks{Subramanya Nageshrao, Ling Zhu, and Dimitar Filev are with Ford Motor Company, Detroit, USA}
\thanks{Contact Email IDs: dhebarya@msu.edu, kdeb@egr.msu.edu}
\thanks{Website: \href{https://coin-lab.org/}{\color{blue}{coin-lab.org}}}
}

\maketitle

\begin{abstract}
Black-box artificial intelligence (AI) induction methods such as deep reinforcement learning (DRL) are increasingly being used to find optimal policies for a given control task. Although policies represented using a black-box AI are capable of efficiently executing the underlying control task and achieving optimal closed-loop performance -- controlling the agent from initial time step until the successful termination of an episode, the developed control rules are often complex and  neither {\em interpretable\/} nor {\em explainable}. 
    In this paper, we use a recently proposed nonlinear decision-tree (NLDT) approach to find a hierarchical set of control rules in an attempt to maximize the open-loop performance for approximating and explaining the pre-trained black-box DRL (oracle) agent using the labelled state-action dataset. Recent advances in nonlinear optimization approaches using evolutionary computation facilitates finding a hierarchical set of nonlinear control rules as a function of state variables using a computationally fast bilevel optimization procedure at each node of the proposed NLDT. 
    %
    Additionally, we propose a re-optimization procedure for enhancing closed-loop performance of an already derived NLDT.
    We evaluate our proposed methodologies (open and closed-loop NLDTs) on  different control problems having multiple discrete actions. In all these problems our proposed approach is able to find relatively simple and interpretable rules involving one to four non-linear terms per rule, while simultaneously achieving on par closed-loop performance when compared to a trained black-box DRL agent. A post-processing approach for simplifying the NLDT is also suggested. The obtained results are inspiring as they suggest the replacement of complicated black-box DRL policies involving thousands of parameters (making them non-interpretable) with relatively simple interpretable policies.
    Results are encouraging and motivating to pursue further applications of proposed approach in solving more complex control tasks.
\end{abstract}


%
\IEEEpeerreviewmaketitle

\input{IEEE_body}

\ifCLASSOPTIONcaptionsoff
  \newpage
\fi



%
\bibliographystyle{IEEEtran}
\bibliography{bibfile}

%

\newpage
\pagebreak

\twocolumn[{%
\centering
    {\huge\bf Supplementary Document}\\[2pt]
\vspace{1cm}
}]

\input{Arxiv_supp_IEEE}

\end{document}

%% file: IEEE_body.tex
\section{Introduction}
Control system problems are increasingly being solved by using modern reinforcement learning (RL) and other machine learning (ML) methods to find an autonomous agent (or controller) to provide an optimal action $A_t$ for every state variable combination $S_t$ in a given environment  at every time step $t$. 
Execution of the output \emph{action} $A_t$ takes the object to the next \emph{state} $S_{t+1}$ in the environment and the process is repeated until a termination criteria is met. The mapping between input \emph{state} $S_t$ and output \emph{action} $A_t$ is usually captured through an artificial intelligence (AI) method. In the RL literature, this mapping is referred to as \emph{policy} $(\pi(S): \mathbb{S} \rightarrow \mathbb{A})$, where $\mathbb{S}$ is the \emph{state space} and $\mathbb{A}$ is the \emph{action space}. Sufficient literature exists in efficient training of these RL \emph{policies} \cite{schulman2015trust, schulman2017proximal, lillicrap2015continuous, mnih2016asynchronous, senda2014acceleration, koga2014stochastic, xu2014clustering}. 
While these methods are efficient at training the AI policies for a given control system task, the developed AI policies, captured through complicated networks, are complex and non-interpretable.

Interpretability of AI policies is important to a human mind due to several reasons: (i) they help provide a better insight and knowledge to the working principles of the derived policies, (ii) they can be easily deployed with a low fidelity hardware, (iii) they may also allow an easier way to extend the control policies for more complex versions of the problem. While defining interpretability is a subjective matter, a number of past efforts have attempted to find interpretable AI policies  with limited success \cite{murdoch2019interpretable, lipton2018mythos}. In this work, we aim at generating policies which are \emph{relatively} interpretable and simple as compared to the black-box AI counterparts such as DNN.

In the remainder of this paper, we first present the main motivation behind finding interpretable policies in Section~\ref{sec:motivation}. A few past studies in arriving at interpretable AI policies 
is presented in Section~\ref{sec:past}. In Section~\ref{sec:nldt}, we review a recently proposed nonlinear decision tree (NLDT) approach in the context of arriving at relatively interpretable AI policies. The overall open-loop and closed-loop NLDT policy generation methods are described in Section~\ref{sec:overall}. Results on four control system problems are presented in Section~\ref{sec:results}. Next, a new benchmark problem is proposed to conduct a scale-up study of our algorithm in Section~\ref{sec:scale_up}. Finally, conclusions and future studies are presented in Section~\ref{sec:conclusions}. Supplementary document provides further details. 

\section{Motivation for the Study}
\label{sec:motivation}
Various data analysis tasks, such as classification, controller design, regression, image processing, etc., are increasingly being solved using artificial intelligence (AI) methods. 
These are done, not because they are new and interesting, but because they have been demonstrated to solve complex data analysis tasks without much change in their usual frameworks. With more such studies over the past few decades \cite{lipton2018mythos}, 
 they are faced with a huge challenge. Achieving a high-accuracy solution does not necessarily satisfy a curious domain expert, particularly if the solution is not interpretable or explainable. 
A technique (whether AI-based or otherwise) to handle data well is no more enough, researchers now demand an explanation of why and how they work \cite{ahmad2018interpretable, goodman2017european}. 

Consider the MountainCar control system problem (see Supplementary document for details), which has been extensively studied using various AI methods \cite{sutton1996generalization, peters2010relative, smart2000practical}. 
The problem has two state variables (position $x_t$ along $x$-axis and velocity $v_t$ along positive $x$-axis) at every time instant $t$ which would describe the state of the car at $t$. Based on the state vector $S_t = (x_t, v_t)$, a policy $\pi(S)$ must decide on one of the three actions $A_t$: decelerate ($A_t=0$) along positive $x$-axis with a pre-defined value $-a$, do nothing ($A_t=1$), or accelerate ($A_t=2$) with $+a$ in  positive $x$-axis direction. The goal of the control policy $\pi(S)$ is to take the under-powered car (it does not have enough fuel to directly climb the mountain and reach the destination) over the right hump in a maximum of 200 time steps starting anywhere at the trough of the landscape. Physical laws of motion are applied and a  policy $\pi(S)$ has been trained to solve the problem. The RL produces a black-box policy $\pi_{oracle}(S)$ for which an action $A_t \in [0,1,2]$ will be produced for a given input $S_t = (x_t, v_t) \in \mathbb{R}^2$. 
\begin{figure}[hbt]
\begin{subfigure}{0.49\linewidth}
    \centering
    \includegraphics[width = 1.07\linewidth]{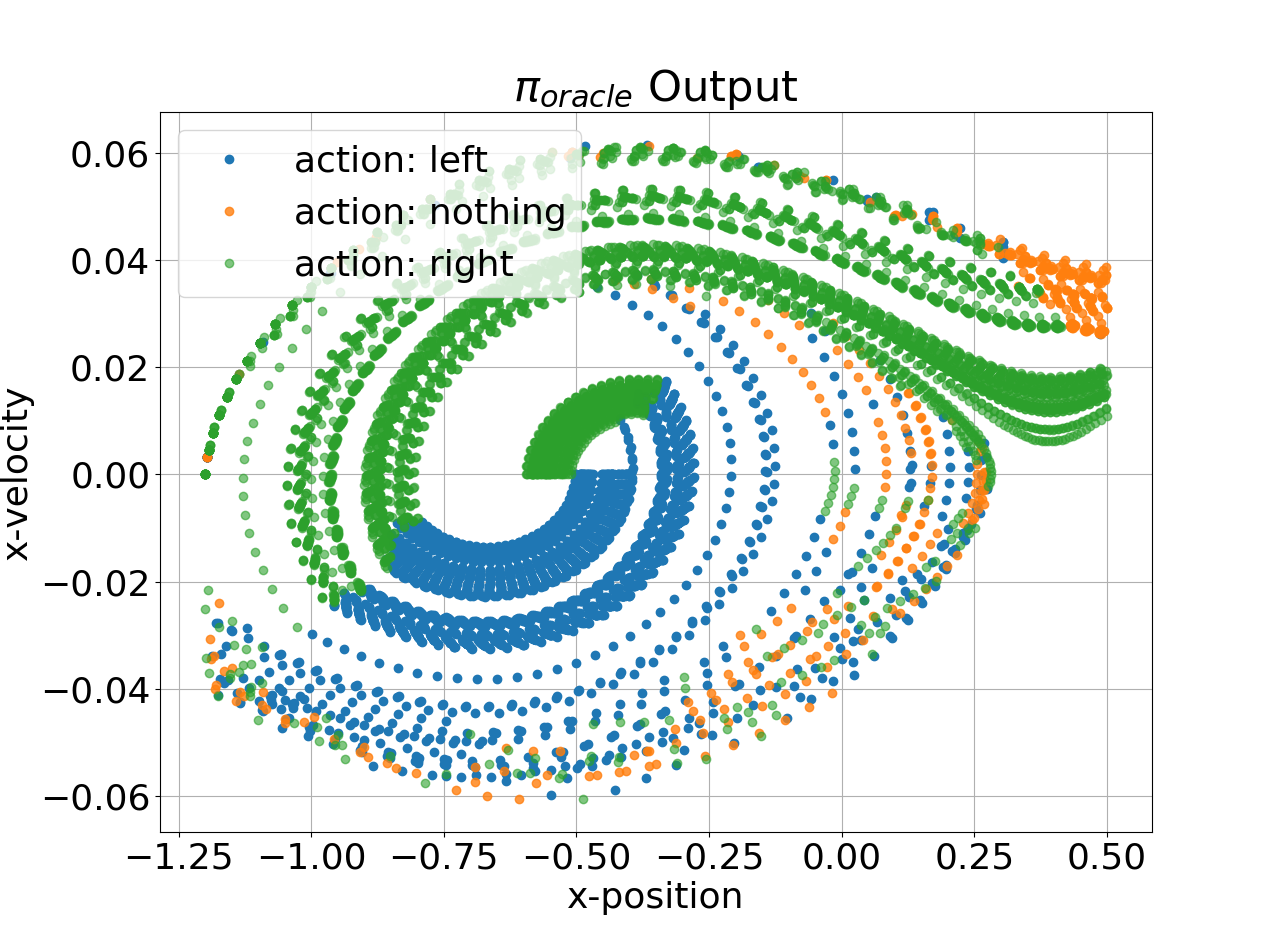}
    \caption{Using $\pi_{oracle}$.}
    \label{fig:m-car-RL}
\end{subfigure}\ 
\begin{subfigure}{0.49\linewidth}
    \centering
    \includegraphics[width = 1.07\linewidth]{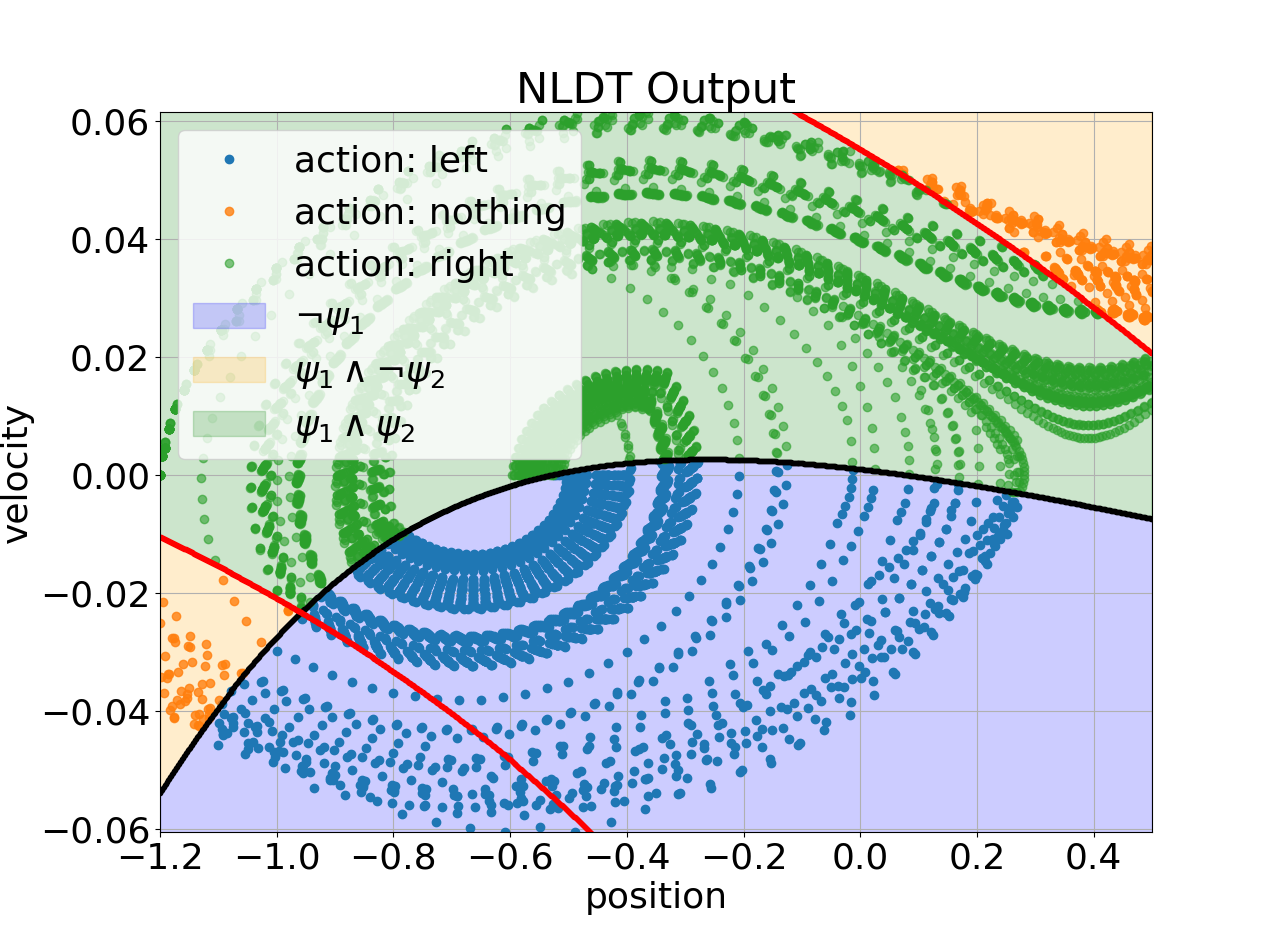}
    \caption{Using NLDT.}
    \label{fig:m-car-NLDT}
\end{subfigure}
\caption{State-action combinations for MountainCar prob.}
\label{fig:mountain_motivation}
\end{figure}
Figure~\ref{fig:m-car-RL} shows the state-action combinations obtained from 92 independent successful trajectories (amounting to total of 10,000 time steps) leading to achieving the goal using a pre-trained deterministic black-box policy $\pi_{oracle}$. 
The $x$-location of the car and its velocity can be obtained from a \emph{point} on the 2D plot. The color of the point $S_t = (x_t, v_t)$ indicates the action $A_t$ suggested by the oracle policy $\pi_{oracle}$ ($A_t=0$: blue, $A_t=1$: orange, and $A_t=2$: green). If a user is now interested in understanding how the policy $\pi_{oracle}$ chooses a correct $A_t$ for a given $S_t$, one way to achieve this would be through an interpretable policy function $\pi_{int}(S_t)$ as follows:
\begin{align}
    \pi_{int}(S_t) &= \left\{\begin{array}{ll}
       0, & \mbox{if $\phi_0(S_t)$ is \emph{true}},\\
       1, & \mbox{if $\phi_1(S_t)$ is \emph{true}},\\
       2, & \mbox{if $\phi_2(S_t)$ is \emph{true}},
       \end{array}\right.
       \label{eq:m-car-rule}
\end{align}
where $\phi_i(S_t): R^2\rightarrow \{0,1\}$ is a Boolean function which partitions the state space $\mathbb{S}$ into two sub-domains based on its output value and for a given state $S_t$, exactly one of $\phi_i(S_t)$ is \emph{true}, thereby making the policy $\pi_{int}$ deterministic.
If we re-look at Figure~\ref{fig:m-car-RL} 
we notice that the three actions are quite mixed at the bottom part of the $x$-$v$ plot (state space). Thus, the partitioning Boolean functions $\phi_i$ need to be quite complex in order to have $\phi_0(S_t)=$~\emph{true} for all \emph{blue} points, $\phi_1(S_t) =$~\emph{true} for all \emph{orange} points and $\phi_2(S_t)=~$\emph{true} for all \emph{green} points.

What we address in this study is an attempt to find an {\em approximated\/} policy function $\pi_{int}(S_t)$ which may not replicate and explain all 100\% time instance data corresponding to the oracle black-box policy $\pi_{oracle}(S_t)$ (Figure~\ref{fig:m-car-RL}), but it is fairly interpretable to explain close to 100\% data. Consider the state-action plot in Figure~\ref{fig:m-car-NLDT}, which is generated with a \emph{relatively} simple and \emph{relatively} interpretable policy $\pi_{int}(S_t) = \{i|\phi_i(S_t)  \text{ is \emph{true}}, i = 1,2,3\}$  obtained by our proposed procedure as shown below:
\begin{equation}
\begin{array} {rcl}
 \phi_0(S_t) &=& \neg\psi_1(S_t), \\
     \phi_1(S_t) &=& \left(\psi_1(S_t) \land \neg\psi_2(S_t)\right), \\
     \phi_2(S_t) &=& \left(\psi_1(S_t) \land \psi_2(S_t)\right),
\end{array}
\label{eq:m_car}
\end{equation}
where 
$\psi_1(S_t) = |0.96-0.63/\widehat{x_t}^2+0.28/\widehat{v_t}-0.22\widehat{x_t}\widehat{v_t}| \leq 0.36$, and 
$\psi_2(S_t) = |1.39-0.28\widehat{x_t}^2-0.30\widehat{v_t}^2| \leq 0.53$ represent black and orange boundaries, respectively. Here, $\widehat{x_t}$ and $\widehat{v_t}$ are normalized state variables (see Supplementary document for details).
The action $A_t$ predicted using the above policy does not match the output of $\pi_{oracle}$ at some states (about 8.1\%), but from our experiments we observe that it is still able to drive the mountain-car to the destination goal located on the right hill in 99.8\% episodes.  

Importantly, the policies are relatively simpler than the corresponding the black-box policy $\pi_{oracle}$ and amenable to an easier understanding of the relationships between $x_t$ and $v_t$ to make a near perfect control. Since the explanation process used the data from $\pi_{oracle}$ as the universal truth, the derived relationships will also provide an explanation of the working of the black-box policy $\pi_{oracle}$. A more gross approximation to Figure~\ref{fig:m-car-RL} by more simplified relationships ($\phi_i$) may reduce the overall \emph{open-loop} accuracy of matching the output of $\pi_{oracle}$. Hence, a balance between a good interpretability and a high \emph{open-loop} accuracy in searching for Boolean functions $\phi_i(S_t)$ becomes an important matter for such an interpretable AI-policy  development study.  

In this paper, we focus on developing a search procedure for arriving at the $\psi$-functions (see Eq.~\ref{eq:m_car}) and their combinations for different discrete action systems. The structure of the policy $\pi_{int}(S_t)$ shown in  Eq.~\ref{eq:m-car-rule} resembles a decision tree (DT), but unlike a standard DT, it involves a nonlinear function at every non-leaf node, requiring an efficient nonlinear optimization method to arrive at reasonably succinct and accurate functionals. 
The procedure we propose here is generic and is independent of the AI method used to develop the black-box policy $\pi_{oracle}$. 

\section{Related Past Studies} \label{sec:past}
\input{literature}

\section{Nonlinear Decision Tree (NLDT) Approach}\label{sec:nldt}
\begin{figure}[hbt]
\centering
\includegraphics[width = 0.9\linewidth]{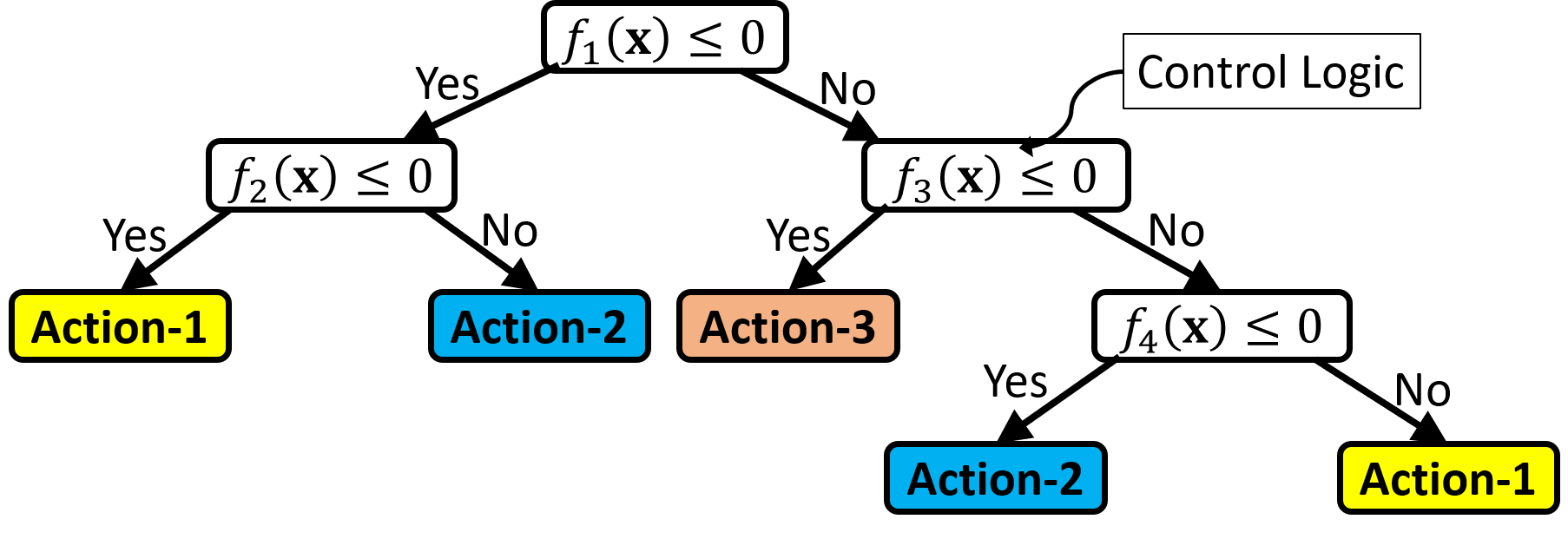}
\caption{Schematic of a binary-split NLDT.}
\label{fig:binary_split_nldt}
\end{figure}

In this study, we use a direct mathematical rule generation approach (presented in Eq.~\ref{eq:m_car}) using a nonlinear decision tree (NLDT) approach \cite{dhebar2020interpretable}, which we briefly describe here. The intention is to model a relatively simple policy $\pi_{int}$ to approximate and explain the pre-trained black-box policy $\pi_{oracle}$ using the labelled state-action data generated using $\pi_{oracle}$.
Decision trees are considered a popular choice due to their interpretability aspects. They are intuitive and each decision can be easily interpreted. However, in a general scenario, regular decision trees often have complicated topology since the rules at each conditional node can assume only axis parallel structure $x_i \le \tau$ to make a split. On the other end, single rule based classifiers like support vector machines (SVMs) have just one rule which is complicated and highly nonlinear. Keeping these two extremes in mind, we develop a nonlinear decision tree framework where each conditional node can assume a nonlinear functional form while the tree is allowed to grow by recursively splitting the data in conditional nodes, similar to the procedure used to induce regular decision trees. In our case of replicating a policy $\pi_{oracle}$, the conditional node captures a nonlinear control logic and the terminal leaf nodes indicate the \emph{action}. This is schematically shown in Figure~\ref{fig:binary_split_nldt}.


In the binary-split NLDT, used in this study, a conditional node is allowed to have exactly two splits as shown in Figure~\ref{fig:binary_split_nldt}. 
The non-linear split rule $f(\mathbf{x})$ 
at each conditional node is expressed as a weighted sum of power laws:
\begin{align}
    f(\mathbf{x}) = \begin{cases}
    \sum_{i=1}^pw_iB_i + \theta_1, \quad \mbox{if $m = 0$},\\
    \left|\sum_{i=1}^pw_iB_i + \theta_1\right| - |\theta_2|, \quad \mbox{if $m = 1$},
    \end{cases}
    \label{eq:binary_split_rule}
\end{align}
where power-laws $B_i$ are given as 
$    B_i = \prod_{j=1}^d x_j^{b_{ij}}$ and $m$   indicates if an absolute operator should be present in the rule or not.
In Section~\ref{sec:nldt_open_loop}, 
we discuss procedures to derive values of exponents $b_{ij}$, weights $w_i$, and biases $\theta_i$.

\section{Overall Approach} \label{sec:overall}
\begin{figure*}[hbt]
    \centering
    \includegraphics[width = 0.9\linewidth]{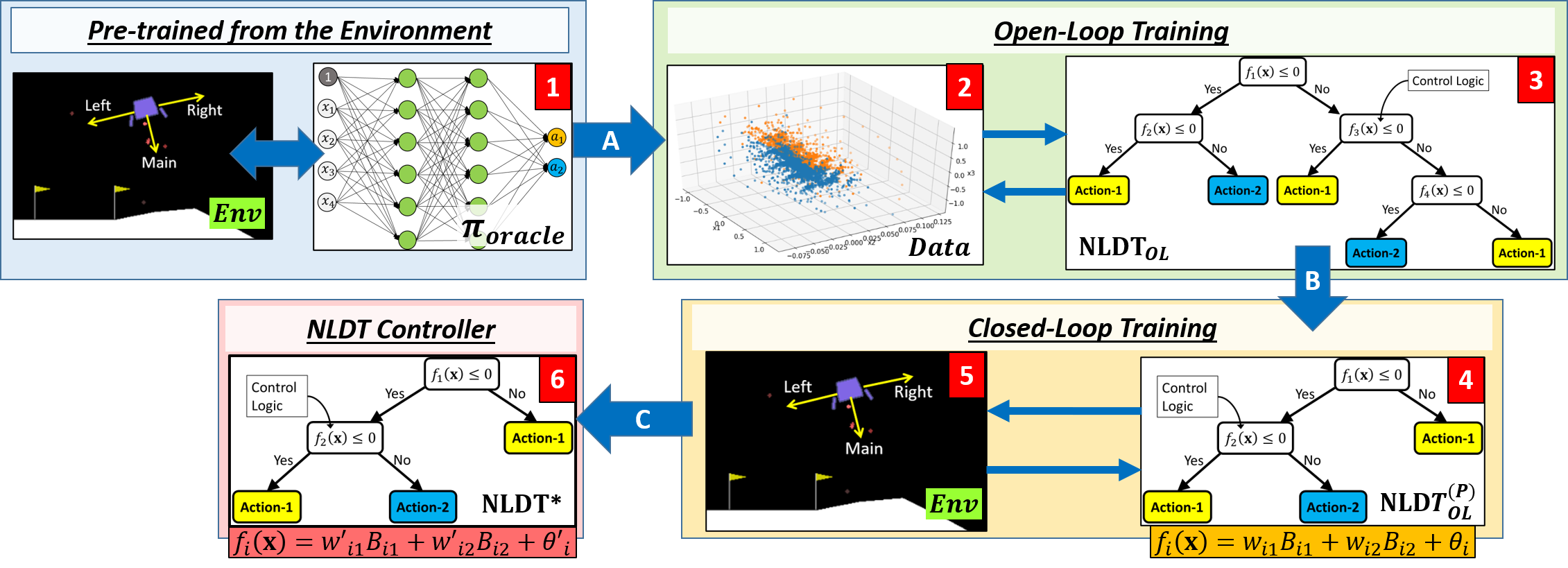}
    \caption{A schematic of the proposed overall approach.}
    \label{fig:iai_rl_flowchart}
\end{figure*}

The overall approach is illustrated in Figure~\ref{fig:iai_rl_flowchart}. 
First, a dedicated black-box policy $\pi_{oracle}$ is trained from the actual environment/physics of the problem. This aspect is not the focus of this paper. Next, the trained policy $\pi_{oracle}$ (Block 1 in the figure) is used to generate labelled training and testing datasets of \emph{state-action} pairs from different time steps. 
We generate two types of training datasets: \emph{Regular} -- as they are recorded from multiple episodes\footnote{an \emph{episode} is one simulation run.}, and \emph{Balanced} -- \emph{selected} from multiple episodes to have almost equal number of \emph{states} for each action, where an \emph{episode} is a complete simulation of controlling an object with a policy over multiple time steps. 
Third, the labelled training dataset (Block 2) is used to find the NLDT (Block 3) using the recursive bilevel evolutionary algorithm described in Section~\ref{sec:nldt_open_loop}. We call this an open-loop NLDT (or, NLDT$_{OL}$), since it is derived from a labelled state-action dataset generated from $\pi_{oracle}$, without using any overall reward or any final goal objective in its search process, which is typically a case while doing reinforcement learning. Use of labelled state-action data in supervised manner allows a faster search of NLDT even with a large dataset as compared to constructing the NLDT from scratch through reinforcement learning by interacting with the environment to maximize the cumulative rewards \cite{verma2018programmatically}.
Next, in an effort to make the overall NLDT interpretable while simultaneously ensuring better closed-loop performance, we prune the NLDT by taking only the top part of NLDT$_{OL}$ (we call NLDT$^{(P)}_{OL}$ in Block 4) and \emph{re-optimize} all non-linear rules within it for the weights and biases using an efficient evolutionary optimization procedure to obtain final NLDT* (Block 5). The re-optimization is done here with closed-loop objectives, such as the cumulative reward function or closed-loop completion rate. 
We briefly discuss the open-loop training procedure of inducing NLDT$_{OL}$ and the closed-loop training procedure to generate NLDT* in next sections.

\subsection{Open-loop Training}
\label{sec:nldt_open_loop}
A labelled state-action dataset is first created using a pre-trained black-box policy $\pi_{oracle}$. Since we are dealing with discrete-action  control problems, the underlying imitation task of replicating the behavior of $\pi_{oracle}$ using the labelled state-action data translates to a \emph{classification} problem. We train NLDT discussed in Section~\ref{sec:nldt} to fit the state-action data through supervised learning.
Nonlinear split-rule $f(\mathbf{x})$ at each conditional node (Figure~\ref{fig:binary_split_nldt} and Eq.~\ref{eq:binary_split_rule}) is derived using a dedicated bilevel optimization algorithm \cite{bilevel-review}, where the upper level searches the \emph{template} of the non-linear rule and the corresponding lower level focuses at estimating optimal values of \emph{weights/coefficients} for optimal split of data present in the conditional node. The optimization formulation for deriving a non-linear split rule $f(\mathbf{x})$ (Eq.~\ref{eq:binary_split_rule}) at a given conditional node 
is given below:
\begin{equation}
\hspace*{-2mm}\begin{array}{rl}
\text{Minimize}&   
    F_U(\mathbf{B},m,\mathbf{w}^{\ast}, \boldsymbol{\theta}^{\ast}), \\
    \text{subject to} & 
 (\mathbf{w}^{\ast}, \boldsymbol{\theta}^{\ast}) \in {\rm argmin}\left\{F_L(\mathbf{w}, \boldsymbol{\theta})|_{(\mathbf{B}, m)} \big| \right. \\
 & \quad F_L(\mathbf{w}, \boldsymbol{\theta})|_{(\mathbf{B},m)} \leq \tau_I,\\ 
 & \quad \left. -1\leq w_i \leq 1, \ \forall i,\ \boldsymbol{\theta} \in [-1,1]^{m+1}\right\},\\
 & m \in \{0,1\}, b_{ij} \in {\cal Z}, 
\end{array}
\label{eq:bilevel_formulation_binary_split}
\end{equation}
where ${\cal Z}$ is a set of exponents allowed to limit the complexity of the derived rule structure. In thus study, we use ${\cal Z}=\{-3,-2,-1,0,1,2,3\}$.
The objective $F_U$ quantifies the complexity of the non-linear rule by enumerating the number of terms present in the equation of the rule $f(\mathbf{x})$ as shown below:
\begin{align}
    \label{eq:simplification_metric}
    & F_U(\mathbf{B},m,\mathbf{w}^{\ast}, \boldsymbol{\theta}^{\ast}) = \sum_{i=1}^{p}\sum_{j=1}^{d}g(b_{ij}),
\end{align}
where $g(\alpha) = 1$, if $\alpha \ne 0$, zero otherwise. $m$ indicates the presence or absence of a modulus operator and $\mathbf{w}$ and $\boldsymbol{\theta}$ encodes rule weights $w_i$ and biases $\theta_i$ respectively. 
The lower level objective function $F_L$ quantifies the net impurity of child nodes resulting from the split. Impurity $I$ of a node $P$ is computed using a Gini-score: 
$\text{Gini}(P) = 1 - \sum_i^c\left(\frac{N_i}{N}\right)^2$,
where $N$ is the total number of points present in the node and $N_i$ represents number of points belonging to class $i$. Datapoints present in node $P$ gets distributed into two non-overlapping subsets based on their split function value. Datapoints with $f(\mathbf{x}) \le 0$ go to the left child node $L$ and rest go to the right child node $R$. The lower level objective function $F_L$ which quantifies the quality of this split is then given by
\begin{equation}
\small
    F_L(\mathbf{w}, \boldsymbol{\theta})|_{(\mathbf{B}, m)} =  \left(\frac{N_L}{N_P}\text{Gini}(L) + \frac{N_R}{N_P}\text{Gini}(R)\right)_{(\mathbf{w},\boldsymbol{\theta},\mathbf{B},m)}.
    \label{eq:F_L_binary_split}
\end{equation}

The $\tau_I$ parameter in Eq.~\ref{eq:bilevel_formulation_binary_split} represents maximum allowable net-impurity (Eq.~\ref{eq:F_L_binary_split}) of child nodes. The resulting child nodes obtained after the split undergo another split and the process continues until one of the termination criteria is met.

More details regarding the bilevel-optimization algorithm \cite{bilevel-review} to derive split-rule $f_i(\mathbf{x})$ at $i$-th conditional node in NLDT can be found in \cite{dhebar2020interpretable}. 

After the entire NLDT is found, in this study, a pruning and tree simplification strategy (see Supplementary Document for more details) is applied to reduce the size of NLDT in an effort to improve on the interpretability of the overall rule-sets. This entire process of inducing NLDT from the labelled state-action data results into the open-loop NLDT -- NLDT$_{OL}$. NLDT$_{OL}$ can then be used to explain the behavior of the oracle policy $\pi_{oracle}$. We will see in Section~\ref{sec:results} that despite being not 100\% accurate in imitating $\pi_{oracle}$, NLDT$_{OL}$ manages to achieve respectable closed-loop performance with 100\% completion rate and a high cumulative reward value. Next, we discuss the closed-loop training procedure to obtain NLDT*.

\subsection{Closed-loop Training}
\label{sec:closed_loop_training}
The intention behind the closed-loop training is to enhance the closed-loop performance of NLDT. It will be discussed in Section~\ref{sec:results} that while closed-loop performance of NLDT$_{OL}$ is at par with $\pi_{oracle}$ on control tasks involving two to three discrete actions, like CartPole and MountainCar, the NLDT$_{OL}$ struggles to autonomously control the agent for control problems such as LunarLander having more states and actions. In closed-loop training, we fine-tune and re-optimize the weights $\mathbf{W}$ and biases $\boldsymbol{\Theta}$ of an entire NLDT$_{OL}$ (or pruned NLDT$_{OL}$, i.e. NLDT$^{(P)}_{OL}$ -- block 4 in Figure~\ref{fig:iai_rl_flowchart}) to maximize its closed-loop fitness ($F_{CL}$), which is expressed as the average of the cumulative reward collected across $M$ episodes:
\begin{equation}
    \begin{array}{rl}
        \text{Maximize} &  F_{CL}(\mathbf{W}, \boldsymbol{\Theta}) = \mathlarger{\frac{1}{M}\sum_{i = 1}^M R_e (\mathbf{W}, \boldsymbol{\Theta})},\\
     \mbox{Subject to}    &  \mathbf{W} \in [-1,1]^{n_w}, \boldsymbol{\Theta} \in [-1,1]^{n_{\theta}},
    \end{array}
    \label{eq:closed_loop}
\end{equation}
where $n_w$ and $n_{\theta}$ are total number of weights and biases appearing in entire NLDT and $M = 20$ in our case.

\section{Results}\label{sec:results}
In this section, we present results obtained by using our approach for control tasks on four problems: (i) CartPole, (ii) CarFollowing, (iii) MountainCar, and (iv) LunarLander. The first two problems have two discrete actions, third problem has three discrete actions, and the fourth problem has four discrete actions. The open-loop statistics are reported using scores of training and testing accuracy on labelled state-action data generated from $\pi_{oracle}$. For quantifying the closed-loop performance, we use two metrics: (i) \emph{Completion Rate} which gives a measure on the number of episodes which are successfully completed, and (ii) \emph{Cumulative Reward}  which quantifies how \emph{well} an episode is executed. 
Fore each problem, 10 runs of open-loop training are executed using 10,000 training datapoints.
 Open-loop statistics obtained from these 10 independent runs of 10,000 training and 10,000 test data each are reported. 
We choose the median performing NLDT$_{OL}$ for closed-loop analysis. We run 50 batches with 100 episodes each and report  statistics of completion-rate and cumulative reward for NLDT* obtained after closed-loop training of median performing NLDT$_{OL}$.

\subsection{CartPole Problem}
This problem comprises of four state variables and two discrete actions. Details regarding this problem are provided in the Supplementary document. The oracle DNN controller is trained using the PPO algorithm \cite{schulman2017proximal}. 
Table~\ref{tab:cartpole_training_data_size} shows the performance of NLDT on training datasets of different sizes. It is observed that NLDT trained with 5,000 and 10,000 data-points shows a robust open-loop performance and also produces $100\%$ closed-loop performance. Keeping this in mind, we keep the training data size of 10,000 fixed across all control problems discussed in this paper. The obtained NLDT$_{OL}$s has about two rules with on an average three terms in the derived policy function.
\begin{table*}[hbt]
\caption{Effect of training data size to approximate performance of NLDT$_{OL}$ on CartPole problem.} 
    \label{tab:cartpole_training_data_size}
    \centering
    \begin{tabular}{|@{\hspace{2pt}}c@{\hspace{2pt}}|c|@{\hspace{2pt}}c@{\hspace{2pt}}|c|c||@{\hspace{3pt}}c@{\hspace{3pt}}|c|}\hline
      \begin{tabular}{c}
          {Training}\\
          {Data Size}
      \end{tabular}   &  \begin{tabular}{c}
          {Training}\\
          {Accuracy}
      \end{tabular} & \begin{tabular}{c}
          {Test. Accuracy}\\
          {(Open-loop)}
      \end{tabular} & 
      \begin{tabular}{c}
           \# \\ Rules
      \end{tabular} & 
      \begin{tabular}{c}
           Rule \\ Length  
      \end{tabular} &
    \begin{tabular}{c}
          {Cumulative}\\
          {Reward, Max=200} 
      \end{tabular} & \begin{tabular}{c}
          {Compl. Rate}\\
          {(Closed-loop)} 
      \end{tabular}\\\hline
      100 & $\mathbf{97.00 \pm  1.55} $ & $ 82.79 \pm 2.40 $ & $1.50 \pm 0.50$ & $\mathbf{3.30 \pm 0.93}$ & $199.73 \pm 0.32$ & $95.00 \pm 5.10$ \\\hline
      500 & $ 95.54 \pm  1.53 $ & $79.66 \pm 3.10$ & $1.90 \pm 0.54$ & $3.88 \pm 0.60$ & $175.38 \pm 2.61$ & $51.00 \pm 5.10$ \\\hline
      1,000 & $ 91.90 \pm  0.87 $ & $ 90.59 \pm 1.87 $ & $1.80 \pm 0.40$ & $4.05 \pm 1.04$ & $\mathbf{200.00 \pm 0.00}$ & $\mathbf{100 \pm 0.00}$ \\\hline
      5,000 & $ 92.07 \pm  1.28 $ & $ 92.02 \pm 1.27 $ & $1.70 \pm 0.46$ & $4.25 \pm 0.90$ & $\mathbf{200.00 \pm 0.00}$ & $\mathbf{100 \pm 0.00}$  \\\hline
      10,000 & $ 91.86 \pm  1.25 $ & $ \mathbf{92.05 \pm 1.10} $ & $\mathbf{1.30 \pm 0.46}$ & $4.45 \pm 1.56$ & $\mathbf{200.00 \pm 0.00}$ & $\mathbf{100 \pm 0.00}$  \\\hline
    \end{tabular}
\end{table*}
Interestingly, the same NLDT (without closed-loop training) also produces $100\%$ closed-loop performance by achieving the maximum cumulative reward value of 200.

\subsubsection{NLDT for CartPole Problem}
\begin{figure}[hbt]
    \centering
    \includegraphics[width = 0.75\linewidth]{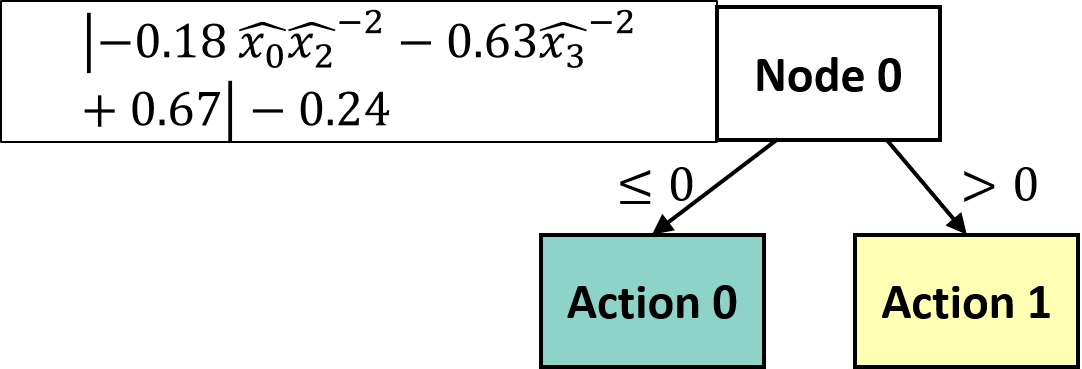}
    \caption{CartPole NLDT$_{OL}$ induced using 10,000 training samples. It is 91.45\% accurate on the testing dataset but has 100\% closed loop performance. Normalization constants are: $\boldx^{\min}$ = [-0.91, -0.43, -0.05, -0.40],
 $\boldx^{\max}$ = [1.37, 0.88, 0.10, 0.45].}
    \label{fig:nldt_CartPole}
\end{figure}

One of the NLDT$_{OL}$ obtained for the CartPole environment is shown in Figure~\ref{fig:nldt_CartPole} in terms of normalized state variable vector $\widehat{\boldx}$. 

The respective policy can be alternatively stated using the programmable \texttt{if-then-else} rule-structure as shown in Algorithm~\ref{algo:nldt_CartPole}:
\begin{algorithm} 
\addtocontents{loa}{\vskip 14.4pt}
\caption{CartPole Rules. Normalization constants are: $x^{min}$ = [-0.91, -0.43, -0.05, -0.40],
$x^{max}$ = [1.37, 0.88, 0.10, 0.45].}
\label{algo:nldt_CartPole}
\eIf{$\left| - 0.18 \widehat{x_{0}}\widehat{x_{2}}^{-2} - 0.63 \widehat{x_{3}}^{-2} + 0.67 \right| - 0.24 \le 0$}
{Action = 0}
{Action = 1}
\end{algorithm}

A little manipulation will reveal that for a correct control startegy, Action~0 must be invoked if following condition is true:
\[2.39\leq \left(\frac{\widehat{x_0}}{\widehat{x_2}^2}+\frac{3.50}{\widehat{x_3}^2}\right) \leq 5.06,\]
otherwise, Action~1 must be invoked. First, notice that the above policy does not require the current velocity ($\widehat{x_1}$) to determine the left or right action movement. Second, for small values of angular position ($\widehat{x_2} \approx 1$) and angular velocity ($\widehat{x_3}\approx 1$), i.e. the pole is falling towards left, the above condition is always true. That is, the cart should be pushed towards left, thereby trying to stabilize the pole to vertical position. On the other hand, if the pole is falling towards right (large values of $\widehat{x_2}\approx 2$ and $\widehat{x_3} \approx 2$), the term in bracket will be smaller than 2.39 for all $\widehat{x_0} \in [1,2]$, and the above policy suggests that Action~1 (push the cart towards right) must be invoked. When the pole is falling right, a push of the cart towards right helps to stabilize the pole towards its vertical position. These extreme case analyses are intuitive and our policy can be explained for its proper working, but what our NLDT approach is able to find is a precise rule for all situations of the state variables to control the Cart-Pole to a stable configuration, mainly using the blackbox-AI data.

\subsection{CarFollowing Problem}
We have developed a discretized version of the car following problem discussed in \cite{nageshrao2019interpretable}. The objective here is to control the rear car and follow a randomly moving car using acceleration or braking actions. This problem involves three state variables and two discrete actions. More details regarding this problem are provided in the Supplementary document. The oracle policy was obtained using a double Q-learning algorithm \cite{van2015deep}. 
The reward function for the CarFollowing problem is shown in the Supplementary document, indicating that a relative distance close to 30$m$ produces the highest reward. 

\begin{figure}[hbt]
    \centering
    \includegraphics[width = 0.9\linewidth]{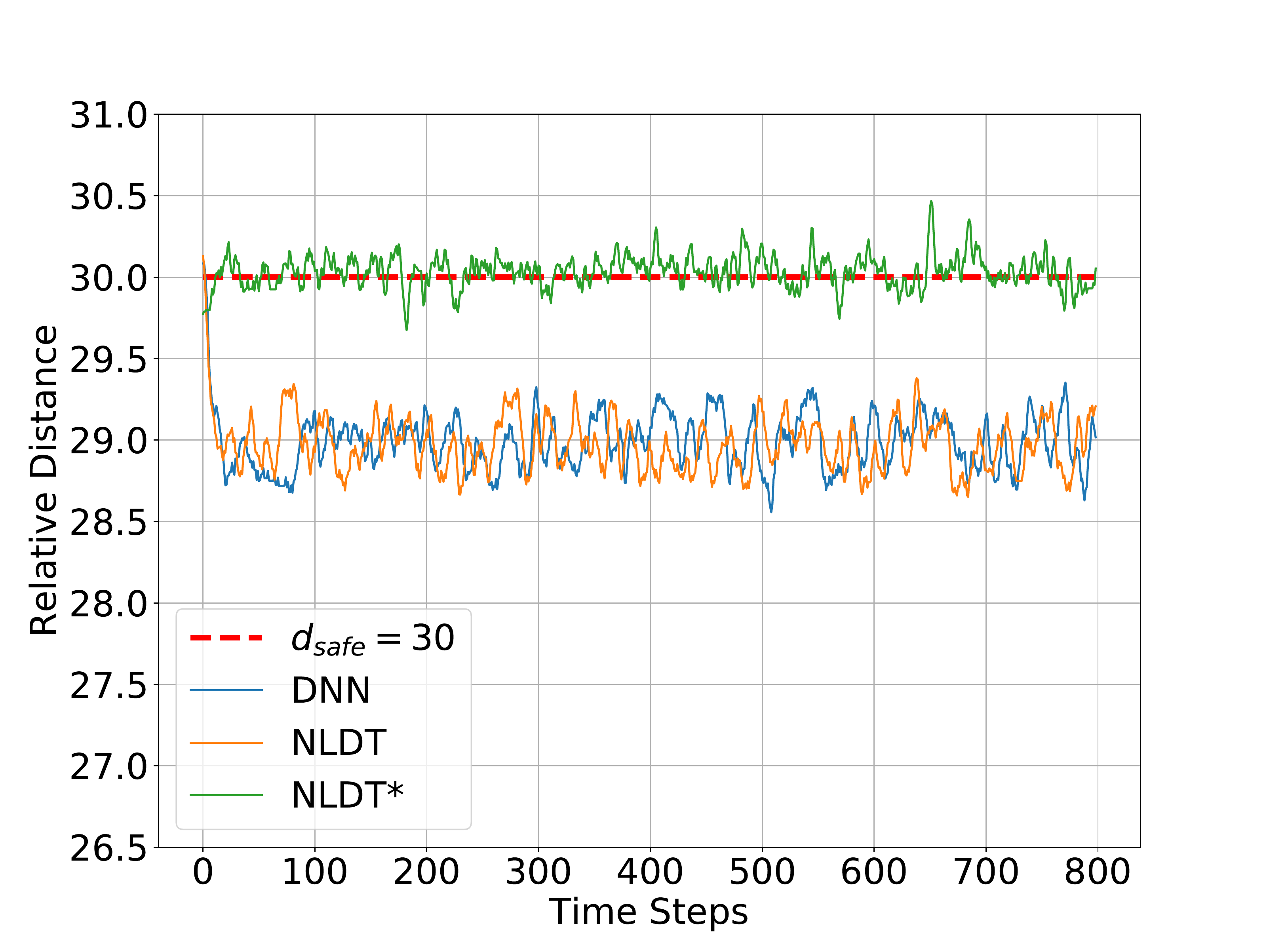}
    \caption{Relative distance plot for CarFollowing.}
    \label{fig:car1d_rel_dist}
\end{figure}

Results for the CarFollowing problem are shown in Table~\ref{tab:nldt_OL_car_following}. An average open-loop accuracy of 96.53\% is achieved with at most three rules, each having 3.28 terms on an average. 
\begin{table}[hbt]
    \centering
    \caption{Results on CarFollowing problem.}
    \label{tab:nldt_OL_car_following}
    \begin{tabular}{|Hp{1cm}|p{1cm}|p{1cm}@{\hspace{2pt}}|Hp{1cm}@{\hspace{2pt}}|p{1cm}@{\hspace{2pt}}||p{1cm}@{\hspace{2pt}}|}\hline
    {Prob.} & {Train. Acc.}
 &  {Test. Acc.}
 & {Depth} & 
         {Total Nodes} & {\# Rules} & 
         {Rule Length} &  
         {Compl. Rate}
\\ \hline
    CarFollowing & $96.41 \pm  1.97 $ & $ 96.53 \pm 1.90 $ & $ 1.90 \pm 0.30 $ & 5 & $2.40 \pm 0.66$ & $3.28 \pm 0.65$ & $100 \pm 0.00$\\ \hline
    \end{tabular}
\end{table}
For this problem, we apply the closed-loop re-optimization (Blocks~4 and 5 to produce Block~6 in Figure~\ref{fig:iai_rl_flowchart}) on the entire NLDT$_{OL}$. As shown Table~\ref{tab:car_following_re_opt}, NLDT* is able to achieve better closed-loop performances (100\% completion rate and slightly better average cumulative reward). 
Figure~\ref{fig:car1d_rel_dist} shows that NLDT* adheres the 30$m$ gap between the cars more closely than original DNN or NLDT$_{OL}$. 
\begin{table}[hbtp]
    \centering
    \caption{Closed-loop performance analysis after re-optimizing NLDT for CarFollowing problem (k $= 10^3$).}
    \label{tab:car_following_re_opt}
    \begin{tabular}{|@{\hspace{2pt}}c@{\hspace{2pt}}|c@{\hspace{2pt}}|c|c@{\hspace{2pt}}|}\hline
   \multirow{2}{*}{AI} & \multicolumn{2}{c|}{Cumulative Reward} & \multirow{2}{*}{Compl. Rate} \\\cline{2-3}
         & Best &  {Avg $\pm$ Std}
        &  \\\hline
        DNN & 174.16k & $173.75$k $ \pm 20.95$ & $\mathbf{100 \pm 0.00}$ \\\hline
        NLDT$_{OL}$ & 174.15k & $173.87$k $ \pm 16.48$ & $\mathbf{100 \pm 0.00}$ \\\hline
        NLDT* & {\bf 179.76k} & $\mathbf{179.71}${\bf k} $\mathbf{ \pm 0.95}$ & $\mathbf{100 \pm 0.00}$\\ \hline
    \end{tabular}
\end{table}

The NLDT corresponding to the CarFollowing problem and its physical interpretation are provided in the Supplementary document.

Results of NLDT's performance on problems with two discrete actions (Tables~\ref{tab:cartpole_training_data_size}, \ref{tab:nldt_OL_car_following} and \ref{tab:car_following_re_opt})
indicate that despite having a noticeable mismatch with the open-loop output of the oracle black-box policy $\pi_{oracle}$, the closed-loop performance of NLDT is at par or at times better than $\pi_{oracle}$. This observation suggests that certain state-action pairs are not of crucial importance when it comes to executing the closed-loop control and, therefore, errors made in predicting these state-action events do not affect and deteriorate the closed-loop performance.

\subsection{MountainCar Problem}
This problem comprises of two state-variables to capture $x$ position and velocity of the car. The task is to use three actions and drive an under-powered car to the destination (see Supplementary Document for more details).

Compilation of results of the NLDT$_{OL}$ induced using training datasets comprising of different data distributions (\emph{regular} and \emph{balanced}) is presented in Table~\ref{tab:mountain_car}. A state-action plot obtained using $\pi_{oracle}$ and one of the NLDT policy corresponding to the first row of Table~\ref{tab:mountain_car} is provided in Figures~\ref{fig:m-car-RL} and \ref{fig:m-car-NLDT}, respectively. It is observed that about $8\%$ mismatch in the open-loop performance (i.e. testing accuracy in Table~\ref{tab:mountain_car}) comes from the lower-left region of state-action plot (Figures~\ref{fig:m-car-RL} and \ref{fig:m-car-NLDT}) due to highly non-linear nature of $\pi_{oracle}$. Despite having this mismatch, the NDLT policy is able to achieve close to $100\%$ closed-loop control performance with an average of 2.4 rules having 2.97 terms.
\begin{table}[hbt]
    \centering
    \caption{Results on MountainCar problem.}
    \label{tab:mountain_car}
    \begin{tabular}{|@{\hspace{2pt}}p{5mm}|@{\hspace{2pt}}p{1cm}|@{\hspace{2pt}}p{1cm}|@{\hspace{2pt}}p{1cm}@{\hspace{2pt}}|@{\hspace{2pt}}Hp{1cm}@{\hspace{2pt}}|@{\hspace{2pt}}p{1cm}@{\hspace{2pt}}||p{1cm}@{\hspace{2pt}}|}\cline{1-7}\cline{8-8}
    {Data}  &  {Train. Acc.} &
         {Test. Acc.} & {Depth} &  
         {Total Nodes} & {\# Rules} & 
         {Rule Length}  &  {Compl. Rate} \\ \cline{1-7}\cline{8-8}
    Reg. & $ {\bf 91.28} \pm  {\bf 0.57} $ & $ {\bf 91.18} \pm {\bf 0.35} $ & $ 2.00 \pm 0.00 $ & 5 & ${\bf 2.40} \pm {\bf 0.49}$ & ${\bf 2.97} \pm {\bf 0.41}$ & $99.00 \pm 1.71$\\ \cline{1-7}\cline{8-8}
    Bal. & $ 81.45 \pm  7.36 $ & $ 87.23 \pm 1.10 $ & $ {\bf 1.90} \pm {\bf 0.30} $ & 9 & $2.80 \pm 0.60$ & $3.07 \pm 0.42$ & $\mathbf{100} \pm \mathbf{0.00}$\\ \cline{1-7}\cline{8-8}
    \end{tabular}
\end{table}
Also, NLDT trained on balanced dataset (2nd row of Table~\ref{tab:mountain_car}) is able to achieve 100\% closed-loop performance and involves about three control rules with an average 1.67 terms in each rule. 

\subsection{LunarLander Problem}
The task in this problem is to control the lunar-lander using four discrete actions and successfully land it on the lunar terrain. The state of the lunar-lander is expressed with eight state variables, of which six are continuous, and two are categorical. More details for this problem are provided in the Supplementary document. 
\begin{table}[hbt]
    \centering
    \caption{NDLT$_{OL}$ with depths 3 and 6 for LunarLander.}
    \label{tab:lunar_lander_result}
    \begin{tabular}{|@{\hspace{2pt}}p{5mm}|@{\hspace{2pt}}c@{\hspace{2pt}}|@{\hspace{1pt}}p{1cm}@{\hspace{5pt}}|@{\hspace{2pt}}p{1cm}@{\hspace{5pt}}|@{\hspace{2pt}}Hp{1cm}@{\hspace{2pt}}|@{\hspace{2pt}}p{1cm}@{\hspace{2pt}}||@{\hspace{2pt}}p{1cm}@{\hspace{5pt}}|}\cline{1-7}\cline{8-8}
{Data}  & {Depth} &  {Train. Acc.} &
         {Test. Acc.}
  &  
         {Total Nodes}
& {\# Rules} & 
         {Rule Length}  &  
         {Compl. Rate} \\ \cline{1-7}\cline{8-8}
    Reg. & 3 & $ 79.17 \pm  1.78 $ & $ 76.36 \pm 3.36 $  & 13 & $5.60 \pm 0.49$ & $5.59 \pm 0.75$ & $14.00 \pm 5.93$ \\\hline 
    Bal. & 3 & $ 69.83 \pm  2.82 $ & $ 66.58 \pm 2.03 $  & 9 & $4.40 \pm 0.66$ & $5.79 \pm 1.31$ & $42.00 \pm 4.40$ \\\hline\hline 
    Reg. & 6 & $ {\bf 87.43} \pm  {\bf 0.65} $ & $ {\bf 81.74} \pm {\bf 0.91} $  & 61 & $34.70 \pm 2.83$ & $4.94 \pm 0.34$ & $48.00 \pm 2.77$ \\\hline 
    Bal. & 6 & $ 81.74 \pm  1.78 $ & $ 71.52 \pm 1.24 $  & 53 & $25.70 \pm 5.83$ & $5.17 \pm 0.37$ & $\mathbf{93.00} \pm \mathbf{3.30}$\\\hline 
    \end{tabular}
\end{table}
\begin{table*}[!hbt]
\caption{Closed-loop performance on LunarLander problem with and without re-optimization on 26-rule NLDT$_{OL}$. 
Number of rules are specified in brackets for each NLDT and total parameters for the DNN is marked.}
    \label{tab:newopt_results}
    \centering
 \begin{tabular}{|@{\hspace{3pt}}c@{\hspace{3pt}}|@{\hspace{3pt}}c@{\hspace{3pt}}|@{\hspace{3pt}}c@{\hspace{3pt}}|@{\hspace{3pt}}c@{\hspace{3pt}}|@{\hspace{3pt}}c@{\hspace{3pt}}|@{\hspace{3pt}}c@{\hspace{3pt}}||c@{\hspace{3pt}}|}\hline
    {Re-Opt.} & {NLDT-2 (2)} & {NLDT-3 (4)} & {NLDT-4 (7)} & {NLDT-5 (13)} & {NLDT-6 (26)} & {DNN (4,996)}\\\hline
    \multicolumn{7}{|c|}{Cumulative Reward}\\\hline
    {Before} & $-1675.77 \pm 164.29$ & $42.96 \pm 13.83$ & $54.24 \pm 27.44$ & $56.16 \pm 23.50$ & $\mathbf{169.43 \pm 23.96}$ & \multirow{2}{*}{$247.27 \pm 3.90$}\\\cline{1-6}
    {After} & $-133.95 \pm 2.51$ & $231.42 \pm 17.95$ & $\mathbf{234.98 \pm 22.25}$ & $182.87 \pm 21.92$ & $214.94 \pm 17.31$ & \\\hline
     \multicolumn{7}{|c|}{Completion Rate}\\\hline
    {Before} & $0.00 \pm 0.00$ & $51.00 \pm 3.26$ & $82.00 \pm 9.80$ & $79.00 \pm 7.66$ & $\mathbf{93.00 \pm 3.30}$ & \multirow{2}{*}{$94.00 \pm 1.96$}\\\cline{1-6}
    {After} & $48.00 \pm 7.38$ & $96.00 \pm 2.77$ & $\mathbf{99.00 \pm 1.71}$ & $93.00 \pm 7.59$ & $94.00 \pm 4.45$ & \\\hline 
    \end{tabular}
\end{table*}

Table~\ref{tab:lunar_lander_result} provides the compilation of results obtained using NLDT$_{OL}$.
In this problem, while a better open-loop performance occurs for regular dataset, a better closed-loop performance is observed when the NLDT open-loop training is done on the balanced dataset. Also, NLDT$_{OL}$ with depth three are not adequate to achieve high closed-loop performance. 
The best performance is observed using balanced dataset where NLDT$_{OL}$ achieves 93\% episode completion rate. A specific NLDT$_{OL}$ with 26 rules each having about 4.15 terms is shown in the Supplementary Document. 

\begin{figure}[hbt]
    \centering
    \includegraphics[width = 0.9\linewidth]{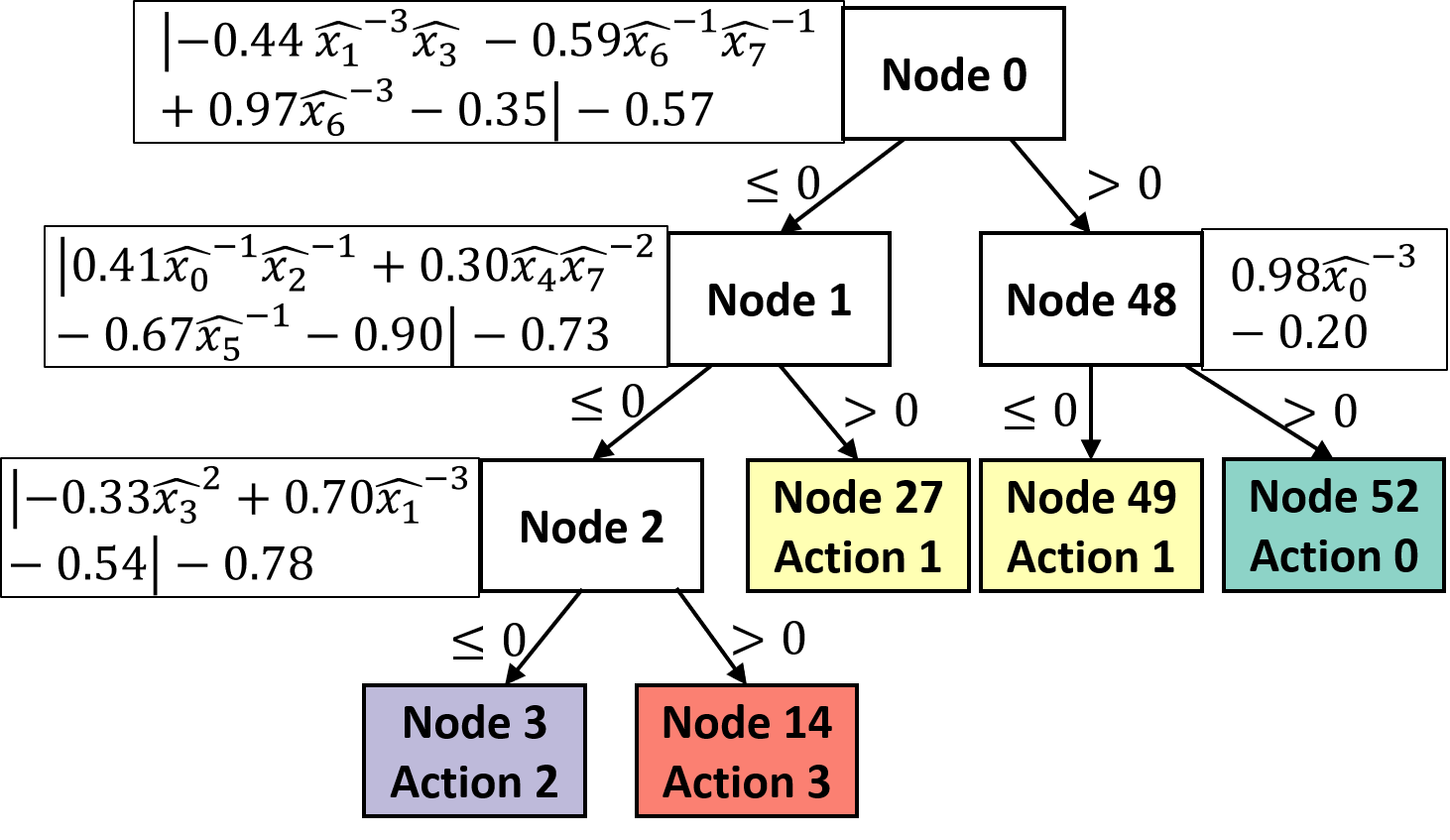}
    \caption{Final NLDT*-3 for LunarLander prob. $\widehat{x_i}$ is a normalized state variable (see Supplementary Document).}
    \label{fig:nldt_3_lunar_lander}
\end{figure}

It is understandable that a complex control task involving many state variables cannot be simplified or made interpretable with just one or two control rules. 
Next, we use a part of the NLDT$_{OL}$ from the root node to obtain the pruned NLDT$^{(P)}_{OL}$ (step `B' in Figure~\ref{fig:iai_rl_flowchart}) and re-optimize all weights ($\mathbf{W}$) and biases ($\boldsymbol{\Theta}$) using the procedure discussed in Section~\ref{sec:closed_loop_training} (shown by orange box in Figure~\ref{fig:iai_rl_flowchart}) to find closed-loop NLDT*. 
Table~\ref{tab:newopt_results} shows that for the pruned NLDT-3 which comprises of the top three layers and involves only four rules of original 26-rule NLDT$_{OL}$ (i.e. NLDT-6), the closed-loop performance increases from 51\% to 96\% (NLDT*-3 results in Table~\ref{tab:newopt_results}) after re-optimizing its weights and biases with closed-loop training. 
The resulting NLDT with its associated four rules are shown in Figure~\ref{fig:nldt_3_lunar_lander}. 

As shown in Table~\ref{tab:newopt_results}, the NLDT* with just two rules (NLDT-2) is too simplistic and does not recover well after re-optimization. However, the
NLDT*s with four and seven rules achieve a near 100\% closed-loop performance. 
Clearly, an NLDT* with more rules (NLDT-5 and NLDT-6) are not worth considering since both closed-loop performances and the size of rule-sets are worse than NLDT*-4. Note that DNN produces a better reward, but not enough completion rate, and the policy is more complex with 4,996 parameters.

\begin{figure}[hbt]
    \centering
    \includegraphics[width = 0.8\linewidth]{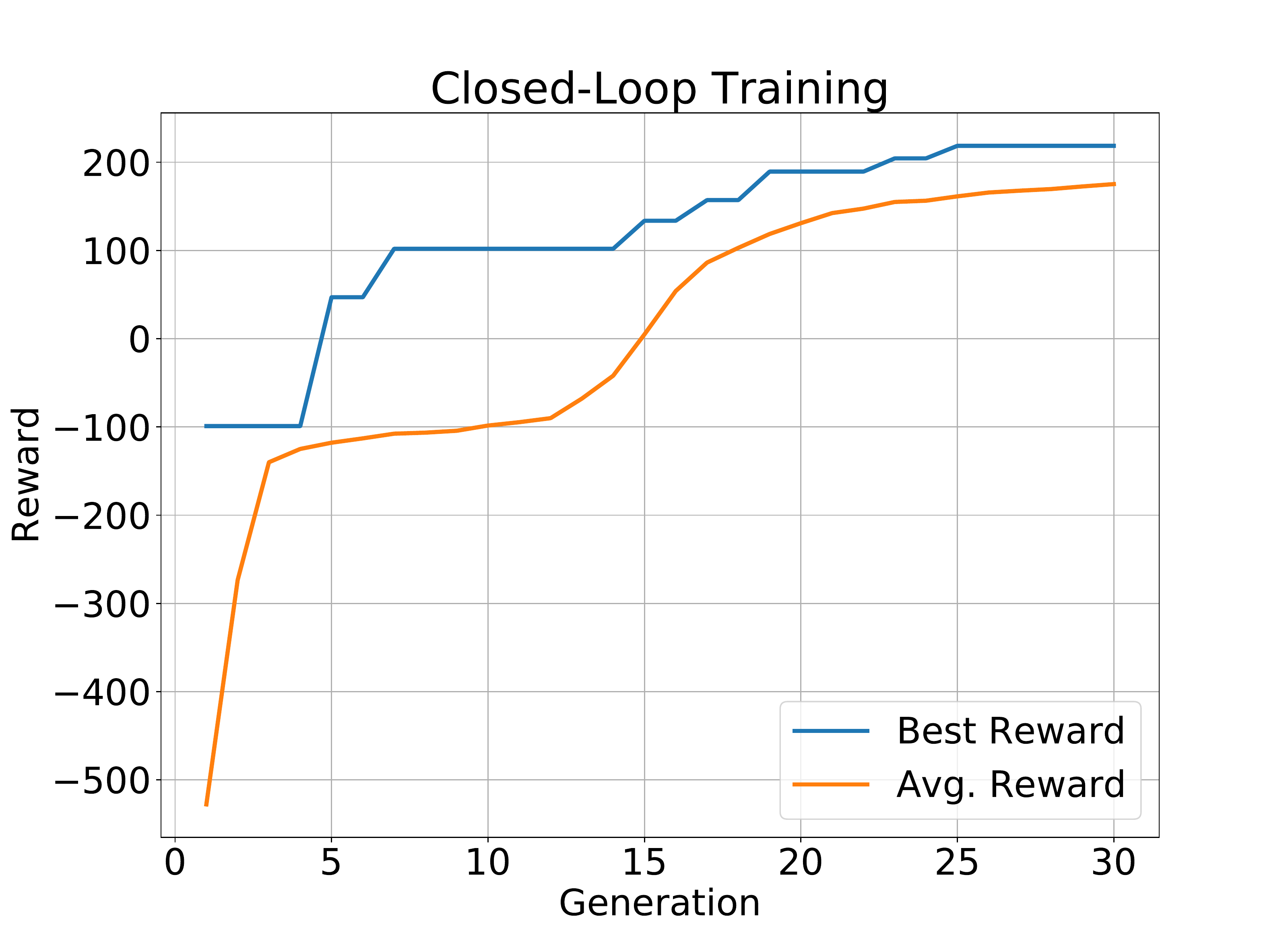}
    \caption{Closed-loop training plot for finetuning the rule-set corresponding to depth-3 NLDT$^{(P)}_{OL}$ to obtain NLDT* for LunarLander problem.}
    \label{fig:closed_loop_training_plot}
\end{figure}

Figure~\ref{fig:closed_loop_training_plot} shows the closed-loop training curve for generating NLDT* from Depth-3 NLDT$^{(P)}_{OL}$. The objective is to maximize the closed-loop fitness (reward) $F_{CL}$ (Eq.~\ref{eq:closed_loop}) which is expressed as the average of the cumulative reward $R_{e}$ collected over $M$ episodes.
It is evident that the cumulative reward for the best-population member climbs to the target reward of 200 at around 25-th generation and the average cumulative reward of the population also catches up the best cumulative reward value with generations.

To check the repeatability of our approach, another run for generating NLDT$_{OL}$ and NLDT* is executed. The resulting NLDT and corresponding equations are provided in the Supplementary document.
A visualization of the real-time closed-loop performance obtained using this new NLDT for two different rule-sets (i.e. before applying re-optimization and after applying the re-optimization through closed-loop training) is shown in \url{https://youtu.be/DByYWTQ6X3E}. It can be observed in the video that the closed-loop control executed using the Depth-3 NLDT$^{(P)}_{OL}$ comprising of rules obtained directly from the open-loop training (i.e. without any re-optimization) is able to bring the LunarLander close to the target. However the LunarLander hovers above the landing pad and the Depth-3 NLDT$^{(P)}_{OL}$ is unable to land it in most occasions. Episodes in these cases are terminated after the flight-time runs out. On the other hand, the Depth-3 NLDT* comprising of rule-sets obtained after re-optimization through closed-loop training is able to successfully land the LunarLander.

\section{Scale-up Study and Improvization on Acrobot Control Problem}
\label{sec:scale_up}
\begin{figure}[hbtp]
    \centering
    \includegraphics[width=0.4\linewidth]{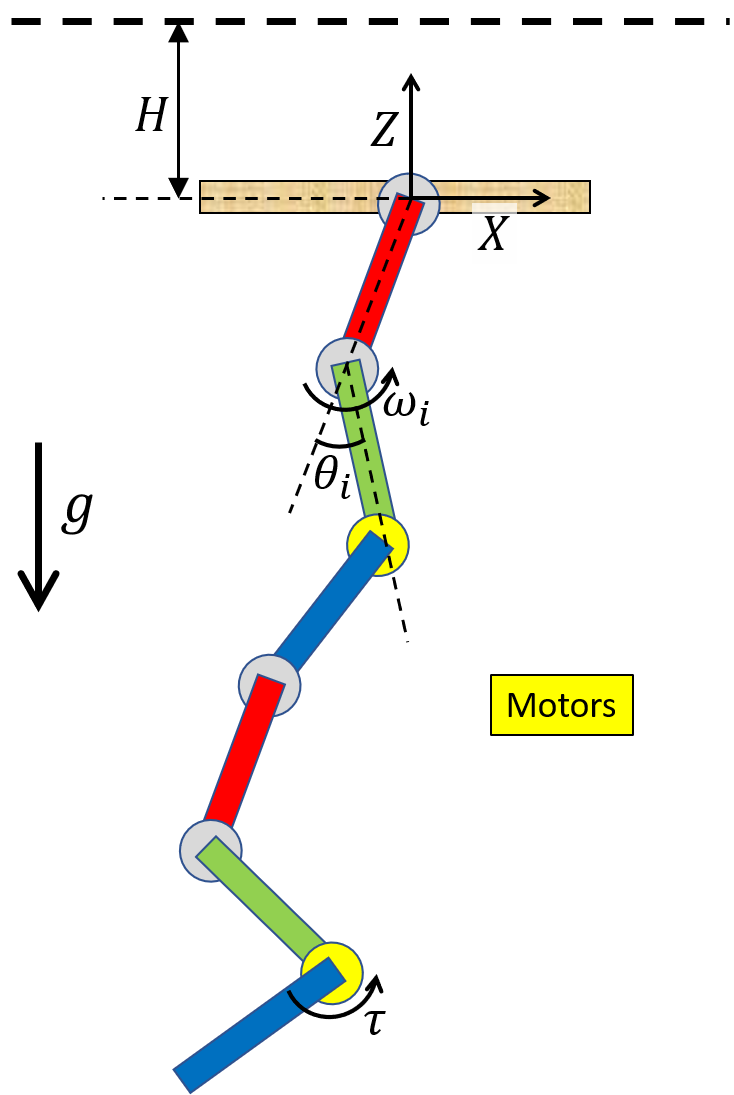}
    \caption{Planar Serial Manipulator (PSM) benchmark problem.}
    \label{fig:planar_serial_m}
\end{figure}

In this section, we investigate how the overall algorithm can be made more efficient in terms of -- \emph{training time} and \emph{scalability}. To this purpose, we introduce a benchmark problem of planar serial manipulator (PSM). This problem is inspired from the classical Acrobot control problem \cite{sutton1996generalization}. A schematic of the PSM problem is provided in Figure~\ref{fig:planar_serial_m}.

The state space for the PSM problem (Figure~\ref{fig:planar_serial_m}) comprises of angular position $\theta_i$ and angular velocity $\omega_i$ of each joint. Thus, for a $n$-link manipulator involving $n$ revolute joints, the state space would be $2n$ dimensional. The motor is located at the last joint of the manipulator and is actuated using three torque values: $-\tau$, $0$ and $\tau$. Each link is of 1 unit length and has its center-of-mass at its geometric center. The motor is assumed to be massless for the sake of simplicity. The base of the manipulator is located at $(0,0,0)$ and the motion of the PSM is limited to the XZ plane. There is a downward gravitational pull ($g$) of 10 units (i.e. -10 along vertical Z axis). Torque is applied along the Y-axis. The task in this problem is to take the end-effector (i.e. tip of the last link, Z-coordinate $z_E$) of the PSM to a desired height of $H$ units by supplying torque to the motor located at the last joint (joint between $(n-1)$-th and $n$-th link). The difficulty of this benchmark problem can be adjusted by
\begin{enumerate}
    \item changing the number of links,
    \item changing the value of desired height level $H$,
    \item changing the value of torque $\tau$, and 
    \item placing extra motors at other joints.
\end{enumerate}
We simulate the mechanics of the planar serial manipulator using \emph{PyBullet} \cite{coumans2020}: a Python based physics engine.

In our work, we provide two case-scenarios by focusing at the first three points of the above list. As mentioned before, by changing the number of links, the dimension of the state-space changes. The dimension of the action-space depends on the number of motors used. In the present work, we keep the number of motors fixed to one, having three discrete actions.

The details regarding two environments which are created and studied in this section are summarized in Table~\ref{tab:manipulator_env_details}.
\begin{table}[hbt]
\caption{Details regarding custom designed Planar Serial Manipulator (PSM) environments.}
    \label{tab:manipulator_env_details}
    \centering
    \begin{tabular}{|c|c|c|c|}\hline
         \textbf{Env. Name} &  \begin{tabular}{c}
              \textbf{Motor} \\ \textbf{Torque} ($\tau$)
         \end{tabular} & \begin{tabular}{c}
              \textbf{Desired} \\ \textbf{Height} ($H$)
         \end{tabular} & \begin{tabular}{c}
              \textbf{\# State} \\ \textbf{Vars.} 
         \end{tabular}\\\hline
         5-Link PSM & 1,000 & $+2$ & 10\\\hline
         10-Link PSM & 2,000 & $+2$ & 20\\\hline
    \end{tabular}
\end{table}

The reward function $r(\mathbf{x},A)$ is given by the following equation:
\begin{align}
    r(\mathbf{x}) = 
    \begin{cases}
    -1 - \left(\frac{H + 1 - z_E}{n + H + 1}\right)^2, \quad \text{ if $z_E < H$,}\\ 100, \quad \text{ if $z_E \geq$ H.}
    \end{cases}
\end{align}
The minimum value for $z_E$ is $-n$ when the entire manipulator is stretched to its full length and all joint angles (i.e. $\theta_i$) are at 0 deg.

At the beginning of an episode, joint angles $\theta_i$ of the manipulator are randomly initialized between $-5$~deg and $+5$~deg, and the angular velocities $\omega_i$ are initialized to a value between $-0.5$~rad/sec and $+0.5$~rad/sec. 

In next sections, we discuss results obtained on the above two custom designed environments by using different procedures of inducing NLDE$_{OL}$. The black-box AI (DNN) is trained using the PPO algorithm \cite{schulman2017proximal}. The resulting DNN has two hidden layers of 64 nodes each and has total 5699 parameters thereby making it massively un-interpretable.

\subsection{Ablation Study for Open-loop Training}
In this section, we launch two separate studies related to open-loop training procedure (see Figure~\ref{fig:iai_rl_flowchart}). It is seen in Section~\ref{sec:nldt_open_loop} that the open-loop training is conducted using a hierarchical bilevel-optimization algorithm, which is discussed at length in \cite{dhebar2020interpretable}. 
A dedicated bilevel-optimization algorithm is invoked to derive the split-rule $f(\mathbf{x})$ at a given conditional node. The upper-level search is executed using a discrete version of a genetic algorithm and the lower-level search is realized through an efficient real-coded genetic algorithm (RGA). Evolutionary algorithms are in general considered robust and have a potential to conduct more global search. However, being population driven, their search-speed is often less than that of classical optimization algorithms. In this section, we study the effect of replacing the real-coded algorithm with a classical \emph{sequential quadratic programming} (SQP) optimization algorithm in the lower-level of the overall bilevel algorithm for obtaining NLDT$_{OL}$. Later, closed-loop training based on real-coded genetic algorithm is applied to NLDT$_{OL}$ to obtain NLDT* by re-optimizing the real valued coefficients of NLDT$_{OL}$ (Section~\ref{sec:closed_loop_training}). We use the \emph{SciPy} \cite{2020SciPy-NMeth} implementation of SQP. The initial point required for this algorithm is obtained using the mixed dipole concept \cite{bobrowski2000induction, kretowski2004evolutionary,krketowski2005global, dhebar2020interpretable}.

For analysis, we induce the NLDT$_{OL}$ of depth-3 on the balanced training dataset of 10,000 datapoints. The testing dataset also comprises of 10,000 datapoints. Comparison of accuracy scores and average training time of inducing NLDT$_{OL}$ by using SQP and RGA algorithm at lower-level is provided in Table~\ref{tab:open_loop_stats}. For a given procedure (SQP or RGA) the best NLDT$_{OL}$ from 10 independent runs is chosen and is re-optimized using closed-loop training (Section~\ref{sec:closed_loop_training}). Statistics regarding closed-loop performance of NLDT* is shown in the last two columns of Table~\ref{tab:open_loop_stats}. It is to note here that the closed-loop training is done using the real-coded genetic algorithm (RGA) discussed in Section~\ref{sec:closed_loop_training}. 

\begin{table*}[hbt]
\caption{Comparing performance of different lower-level optimization algorithms. For comparison, closed-loop performance of the original DNN policy is also reported.}
    \label{tab:open_loop_stats}
    \centering
    \begin{tabular}{|@{\hspace{1pt}}c@{\hspace{1pt}}|c|c|c||c|@{\hspace{2pt}}c@{\hspace{2pt}}|}\hline
     & \multicolumn{3}{c||}{\textbf{Open-Loop} NLDT$_{OL}$} & \multicolumn{2}{c|}{\textbf{Closed-Loop} NLDT*}\\\hline
       \begin{tabular}{c}
            \textbf{Algo.} \\ \textbf{Name}
       \end{tabular}  &  
       \begin{tabular}{c}
            \textbf{Training} \\ \textbf{Accuracy}
       \end{tabular} & 
       \begin{tabular}{c}
            \textbf{Testing} \\ \textbf{Accuracy} 
       \end{tabular} & 
       \begin{tabular}{c}
            \textbf{Training} \\ \textbf{Time (s)}
       \end{tabular} & 
       \begin{tabular}{c}
            \textbf{Cumulative} \\ \textbf{Reward}
       \end{tabular} & 
       \begin{tabular}{c}
            \textbf{Completion} \\ \textbf{Rate}
       \end{tabular}\\\hline
       \multicolumn{6}{|c|}{5-Link Manipulator} \\\hline
        SQP & $62.46 \pm 2.01$ & $\mathbf{69.34 \pm 5.39}$ & $\mathbf{15.29 \pm 4.95}$ & $\mathbf{-146.81 \pm 8.53}$ & $\mathbf{96.00 \pm 2.77}$ \\\hline
        RGA & $\mathbf{71.14 \pm 1.77}$ & $69.17 \pm 4.39$ & $1091.56 \pm 319.18$ & $-152.64 \pm 6.62$ & $99.00 \pm 1.71$ \\\hline
        \textit{DNN} & NA & NA & NA & $-183.35 \pm 12.22$ & $89.00 \pm 5.19$ \\\hline\hline
        \multicolumn{6}{|c|}{10-Link Manipulator}\\\hline
        SQP & $56.57 \pm 1.00$ & $55.64 \pm 3.58$ & $\mathbf{39.27 \pm 11.63}$ & $-318.82 \pm 15.52$ & $\mathbf{96.00 \pm 3.92}$\\\hline
        RGA & $\mathbf{65.84 \pm 0.85}$ & $\mathbf{62.60 \pm 3.09}$ & $2860.96 \pm 789.49$ & $\mathbf{-281.88 \pm 9.38}$ & $95.00 \pm 4.31$ \\\hline
        \textit{DNN} & NA & NA & NA & $-325.86 \pm 4.63$ & $85.88 \pm 1.94$ \\\hline
    \end{tabular}
\end{table*}

It can be observed from the results that open-loop training done with SQP in lower-level is about 70 times faster than the training done using RGA at the lower level. However, the training done with RGA has a better overall open-loop performance. Thus, if the task is to closely mimic the behavior of black-box AI or if only a high classification accuracy is desired (in case of classification problems), then RGA is the recommended algorithm for lower-level optimization to obtain NLDT$_{OL}$. However,  NLDT* obtained after re-optimizing NLDT$_{OL}$ corresponding to SQP and RGA have similar closed-loop completion rate (last column of Table~\ref{tab:open_loop_stats}). This implies that despite low open-loop accuracy scores, the open-loop training done using SQP in lower-level is able to successfully determine the template of split-rules $f(\mathbf{x})$ and the topology of NLDT, which upon re-optimization via closed-loop training algorithm can fetch a decent performing NLDT*. During open-loop training, the search on weights and coefficients using SQP is possibly not as perfect as compared to the one obtained through RGA, however, the re-optimization done through closed-loop training can compensate this shortcoming of SQP algorithm and produce NLDT* with a respectable closed-loop performance. Additionally, in either cases, the NLDT* obtained always has a better closed-loop performance than the original black-box DNN policy. This observation suggests that it is preferable to use SQP in lower-level during open-loop training to quickly arrive at a rough structure of NLDT$_{OL}$ and then use closed-loop training to derive a high performing NLDT*. 

\subsection{Closed-loop Visualization}
In this section, we provide a visual insight into the closed-loop performance of DNN and NLDT* which we derived in the previous section. In our case, the frequency of the simulation is set to 240Hz, meaning that the transition to the next state is calculated using the time-step of 1/240 seconds. Geometrically speaking, this implies that the Euclidean distance between states from neighboring time-steps would be small. The AI (DNN or NLDT*) outputs the action value of 0 ($-\tau$ torque), 1 ($0$ torque) or 2 ($+\tau$ torque) for a given input state. Action Vs Time plots corresponding to different closed-loop simulation runs obtained by using DNN, NLDT* (SQP)\footnote{NLDT* (SQP) indicates that the corresponding NLDT$_{OL}$ was derived using the SQP algorithm in the lower-level} and NLDT* (RGA)\footnote{NLDT* (RGA) indicates that the corresponding NLDT$_{OL}$ was derived using the RGA algorithm in the lower-level} as controllers is shown in Figure~\ref{fig:5_link_plots}  for 5-link manipulator problem (plots for 10-link PSM are provided in the Supplementary document).
\begin{figure*}[hbtp]
    \centering
    \begin{subfigure}[b]{0.32\linewidth}
         \centering
         \includegraphics[width=\linewidth]{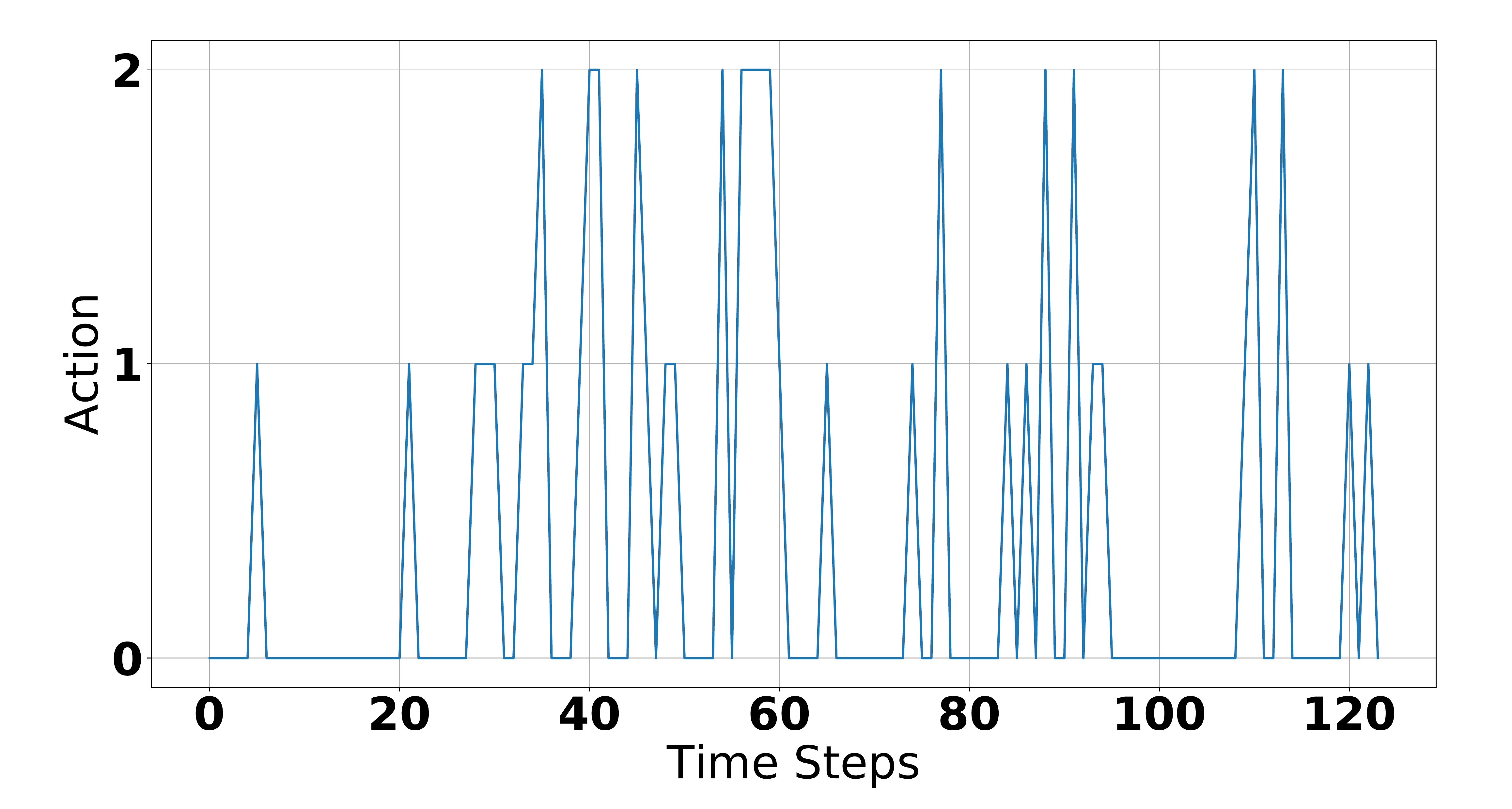}
         \caption{DNN}
         \label{fig:5_link_DNN}
     \end{subfigure}
     \hfill
     \begin{subfigure}[b]{0.32\linewidth}
         \centering
         \includegraphics[width=\linewidth]{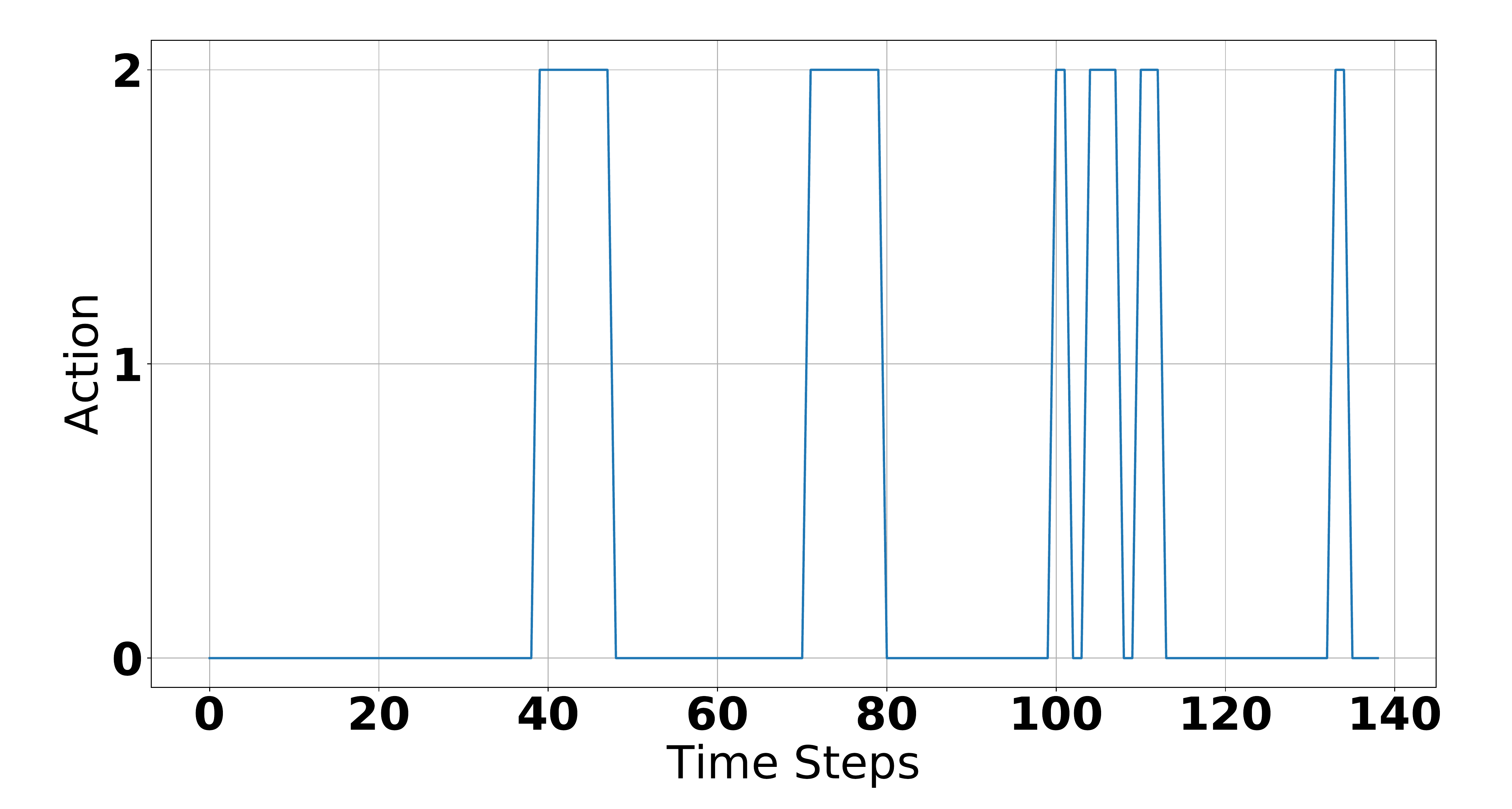}
         \caption{NLDT* (SQP)}
         \label{fig:5_link_SQP}
     \end{subfigure}
     \hfill
     \begin{subfigure}[b]{0.32\linewidth}
         \centering
         \includegraphics[width=\linewidth]{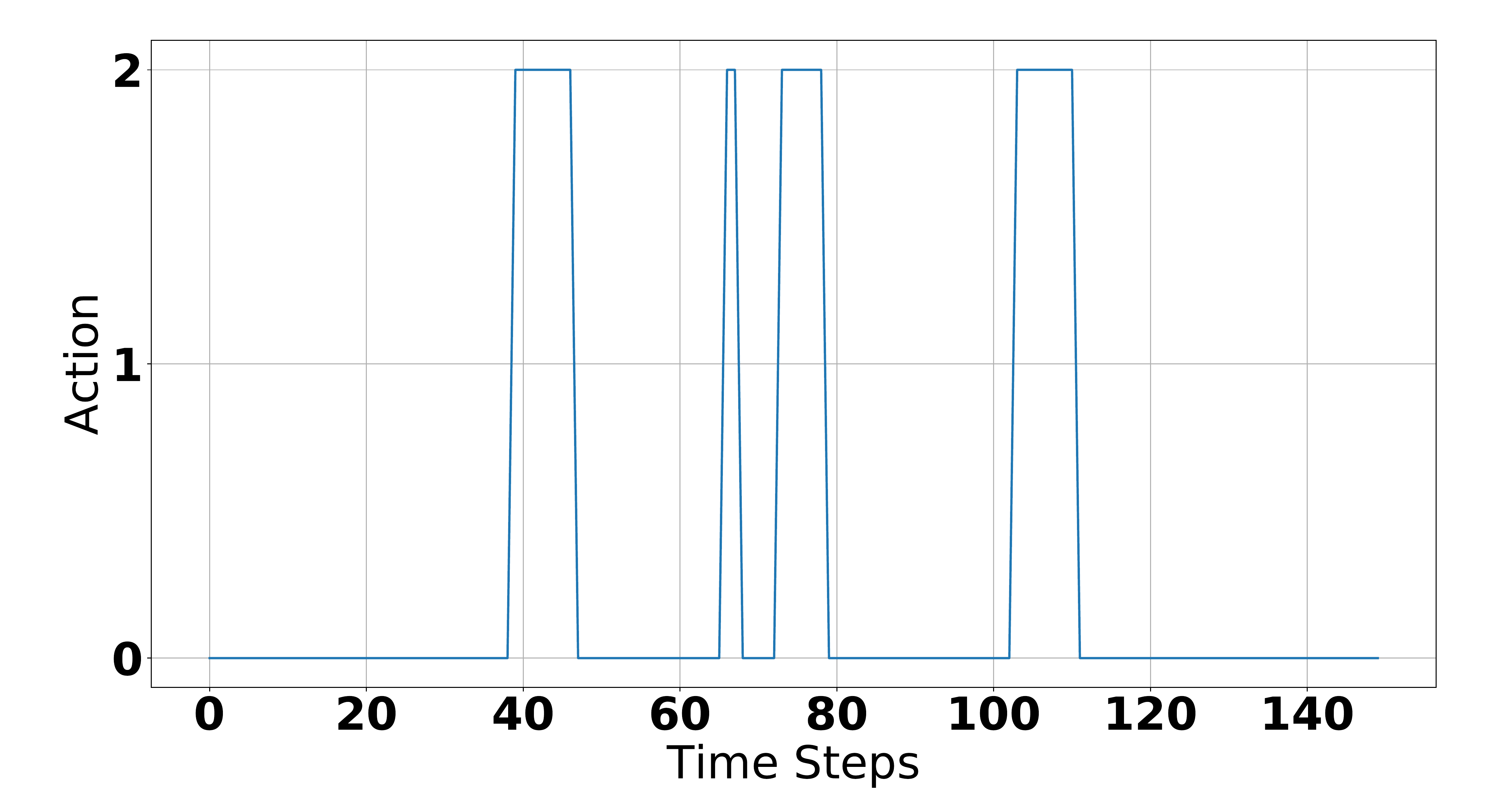}
         \caption{NLDT* (RGA)}
         \label{fig:5_link_RGA}
     \end{subfigure}
    \caption{Action Vs. Time plot for 5-Link manipulator problem. Figure~\ref{fig:5_link_SQP} provides the plot for NLDT* which is obtained from the NLDT$_{OL}$ trained using SQP algorithm in lower-level. Similarly, Figure~\ref{fig:5_link_RGA} provides the plot for NLDT* which is obtained from the NLDT$_{OL}$ trained using RGA algorithm in lower-level.}
    \label{fig:5_link_plots}
\end{figure*}

Certain key observations can be made by looking at the plots in Figure~\ref{fig:5_link_plots}. The control output for DNN is more erratic, with sudden jerks as compared to the control output of NLDT*~(SQP) and NLDT*~(RGA). The performance of NLDT* in Figures~\ref{fig:5_link_SQP} and \ref{fig:5_link_RGA} is smooth and regular. This behaviour can be due to the involvement of a relatively simpler non-linear rule set (as compared to the complicated non-linear rule represented by DNN) which are captured inside NLDT*. This is equivalent to the observation we made for the mountain car problem in Figures~\ref{fig:m-car-RL} and \ref{fig:m-car-NLDT}, wherein the black-box AI had a very erratic behavior for the region of state-space in the lower-half of the state-action plot, while the output of NLDT was more smooth. Additionally, it was seen in Table~\ref{tab:open_loop_stats} that the NLDT* (irrespective of how its predecessor NLDT$_{OL}$ was obtained, i.e. either through SQP or RGA in lower-level) showed better closed-loop performance than the parent DNN policy. This observation implies that simpler rules expressed in the form of a nonlinear decision tree have better generalizability, thereby giving more robust performance for randomly initialized control problems. A careful investigation to the plots in Figure~\ref{fig:5_link_SQP} and \ref{fig:5_link_RGA} reveals that only two out of three allowable actions are required to efficiently execute the given control task of lifting the end-effector of a 5-link serial manipulator. This concept will be used to re-engineer the NLDT*, a discussion regarding which is provided in the next section.


\subsection{Reengineering NLDT*}
\label{sec:re_engineering_nldt}
\begin{figure}[hbt]
    \centering
    \begin{subfigure}[b]{\linewidth}
        \centering
        \includegraphics[width = \linewidth]{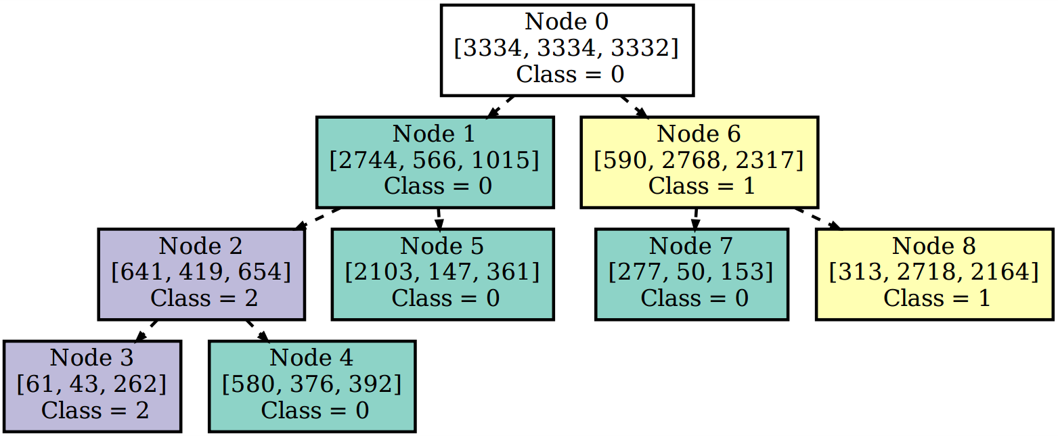}
        \caption{NLDT$_{OL}$}
        \label{fig:5_link_NLDT_OL}
    \end{subfigure}%
    \vspace{5pt}
    \begin{subfigure}[b]{0.85\linewidth}
        \centering
        \includegraphics[width = \linewidth]{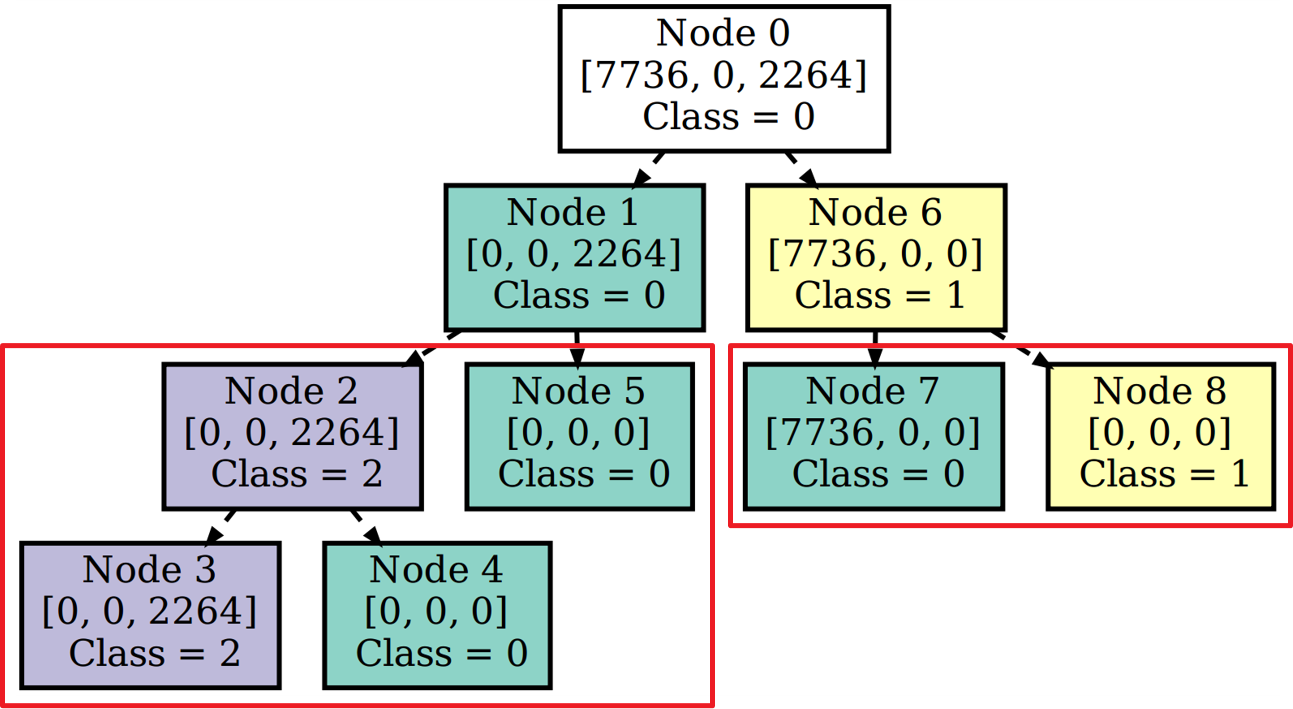}
        \caption{NLDT*}
        \label{fig:5_link_NLDT_CL}
    \end{subfigure}%
    \caption{NLDTs for 5-Link Manipulator problem.}
    \label{fig:nldt_re_engineering}
\end{figure}

It is seen in action-time plots in Figure~\ref{fig:5_link_SQP} and Figure~\ref{fig:5_link_RGA} that not all actions are required to perform a given control task. Also, it may be possible that while performing a closed-loop control using NLDT, not all branches and nodes of NLDT are visited. Thus, the portion of the NLDT which is not being utilized or is getting utilized very rarely can be pruned and the overall NLDT architecture can be made simpler. To illustrate this idea, we consider the NLDT which is derived for the 5-link manipulator problem. The topology of the best performing NLDT$_{OL}$ (SQP) for the 5-link manipulator problem is shown in  Figure~\ref{fig:5_link_NLDT_OL}. 

As mentioned before, this NLDT$_{OL}$ is trained on a balanced training dataset which is generated by collecting state-action pairs using the oracle DNN controller. In the figure, for each node, the information regarding its node-id, class distribution (given in square parenthesis) and the most dominating class is provided. Other than the root-node (Node 0), all nodes are colored to indicate the dominating class, however, it is to note that only the class associated to leaf-nodes carry the actual meaning while predicting the action for a given input state. This NLDT$_{OL}$ comprises of four split-rules in total. The class-distribution for each node is obtained by counting how many datapoints from the balanced training dataset visited a given node. Thus, the root node comprises of all datapoints (total 10,000), which are then scattered according to the split-rules present at each conditional node.

\begin{figure}[htb]
    \centering
    \includegraphics[width = 0.7\linewidth]{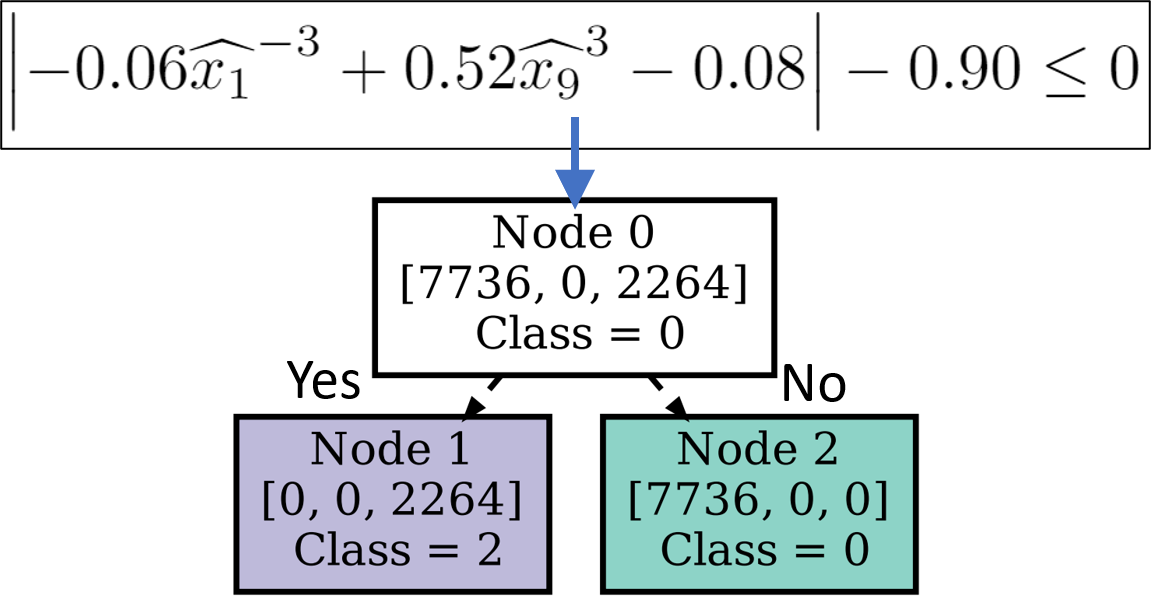}
        \caption{Pruned version of NLDT* (Figure~\ref{fig:5_link_NLDT_CL}) for 5-link manipulator problem.}
        \label{fig:5_link_NLDT_CL_pruned}
\end{figure}

Figure~\ref{fig:5_link_NLDT_CL} provides topology of NLDT* which is obtained after re-optimizing NLDT$_{OL}$ of Figure~\ref{fig:5_link_NLDT_OL} using closed-loop training. As discussed in Sections~\ref{sec:nldt_open_loop} and \ref{sec:closed_loop_training}, the topology of the tree and the structure of non-linear rules is identical for both: NLDT$_{OL}$ and NDLT*. However, the weights and biases of NLDT* are updated to enhance the closed-loop control performance. Similar to NLDT$_{OL}$ of Figure~\ref{fig:5_link_NLDT_OL}, the information regarding node id and class-distribution is provided for all the nodes of NLDT* in Figure~\ref{fig:5_link_NLDT_CL}. However, the data distribution in NLDT* is obtained by using the actual state-action data from closed-loop simulations, wherein NLDT* is used as a controller. Total 10,000 datapoints are collected in form of sequential states-action pairs from closed-loop simulation runs which are executed using NLDT*. As can be seen in the root node of NLDT* (Figure~\ref{fig:5_link_NLDT_CL}), out of 10,000 states visited during closed-loop control, action 0 ($-\tau$ torque) was chosen by NLDT* for total 7,736 states and action 2 ($+\tau$ torque) was chosen for 2,264 states. In none of the states visited during closed-loop control was action 1 (no torque) chosen. This is consistent with  what we have observed in the Action Vs Time plot in Figure~\ref{fig:5_link_SQP}, wherein most of the time action 0 was executed, while there was no event where action 1 was executed. The flow of these 10,000 state-action pairs through NLDT* and their corresponding distribution in each node of NLDT* is provided in Figure~\ref{fig:5_link_NLDT_CL}. It can be observed that Node~5, Node~4 and Node~8 of NLDT* are never visited during closed-loop control. This implies that splits at Node~2, Node~1 and Node~6 are redundant. Thus, the part of NLDT* shown in red-box in Figure~\ref{fig:5_link_NLDT_CL} can be pruned and the overall topology of the tree can be simplified. The pruned NLDT* will involve only one split (occurring at Node~0) and two leaf nodes: Node~1 and Node~6. However, it is to note here that we need to re-assign class-labels to the newly formed leaf nodes (i.e. Node~1 and Node~6) based on the data-distribution from closed-loop simulations. The old class-labelling for the Node~1 and Node~6 was done based on the open-loop data (Figure~\ref{fig:5_link_NLDT_OL}). Using the new class distribution corresponding to NLDT* (Figure~\ref{fig:5_link_NLDT_CL}), Node~1 is re-labelled with Class-2 and Node~6 with Class-0. The pruned version of NLDT* of Figure~\ref{fig:5_link_NLDT_CL} is provided in Figure~\ref{fig:5_link_NLDT_CL_pruned} (here nodes are re-numbered, with Node~6 of NLDT* in Figure~\ref{fig:5_link_NLDT_CL} re-numbered to Node~2 in the pruned NLDT* as shown in Figure~\ref{fig:5_link_NLDT_CL_pruned}). 
The split-rule corresponding to the root-node is also shown. Interestingly, out of 10 total state-variables, only 2 are used to decide which action to execute for closed-loop control: $x_1$ corresponds to the angular position of the second link and $x_9$ variable corresponds to the angular velocity of the last joint.
The above rule indicates a single and interpretable relationship: action~2 must be invoked when  $\widehat{x}_9 \leq \sqrt[3]{\left(1.885 - 0.115/\widehat{x}_1^3\right)}$, otherwise action~0 must be invoked. 
This NLDT* corresponds to a cumulative reward of $-139.78 \pm 7.65$ and performs at $99.00 \pm 1.71$\% completion rate, which provides a remarkably simple interpretation of the control strategy for this apparently complex problem. 


\section{Conclusions}\label{sec:conclusions}
  In this paper, we have proposed a two-step strategy to arrive at hierarchical and relatively interpretable rulesets using a nonlinear decision tree (NLDT) concept to facilitate an explanation of the working principles of AI-based policies. The NLDT training phases use recent advances in nonlinear optimization to focus its search on rule structure and details describing weights and biases of the rules by using a bilevel optimization algorithm. Starting with an open-loop training, which is relatively fast but uses only time-instant state-action data, we have proposed a final closed-loop training phase in which the complete or a part of the open-loop NLDT is re-optimized for weights and biases using complete episode data. Results on four popular discrete action problems have amply demonstrated the usefulness of the proposed overall approach. 
  
  This proof-of-principle study encourages us to pursue a number of further studies. First, the scalability of the NLDT approach for challenging large-dimensional state-action space problems must now be explored. A previous study on NLDT \cite{dhebar2020interpretable} on binary classification of dominated versus non-dominated data in multi-objective problems was successfully extended to 500-variable problems. While it is encouraging, the use of customization methods for initialization and genetic operators using problem heuristics and/or recently proposed {\em innovization\/} methods \cite{deb2006innovization} in the upper level problem can be tried. Second, this study has used a computationally fast open-loop accuracy measure as the fitness for evolution of the NLDT$_{OL}$. This is because, in general, an NLDT$_{OL}$ with a high open-loop accuracy is likely to achieve a high closed-loop performance. However, we have observed here that a high closed-loop performance is achievable with a NLDT$_{OL}$ having somewhat degraded open-loop performance, but re-optimized using closed-loop performance metrics. 
  Thus, a method to identify the crucial (open-loop) states from the AI-based controller dataset that improves the closed-loop performance would be another interesting step for deriving NLDT$_{OL}$. This may eliminate the need for re-optimization through closed-loop training. Third, a more comprehensive study using closed-loop performance and respective complexity as two conflicting objectives for a bi-objective NLDT search would produce multiple trade-off control rule-sets. Such a study can, not only make the whole search process faster due to the expected similarities among multiple policies, they will also enable users to choose a single policy solution from a set of accuracy-complexity trade-off solutions. 
  

%% file: literature.tex
In \cite{noothigattu2018interpretable}, an interpretable orchestrator is developed to choose from two RL-policies $\pi_C$ for maximizing reward and $\pi_R$ for maximizing an {\em ethical\/} consideration.
The orchestrator is dependent on only one of the state-variables and despite it being interpretable, the policies: $\pi_C$ and $\pi_R$ are still black-box and convoluted. 
\cite{maes2012policy} constructs a set of interpretable index based policies and uses multi-arm bandit procedure to select a high performing index based policy. The search space of interepretable policies is much smaller and the procedure suggested for finding an interpretable policy is computationally heavy, taking about hours to several days of computational time on simple control problems.
In \cite{hein2018interpretable}, genetic programming (GP) is used to obtain interpretable policies on control tasks involving continuous actions space through model-based policy learning. However the \emph{interpretability} was not captured in the design of the fitness function and a large archive was created passively to store every policy for each complexity encountered during the evolutionary search. A linear decision tree (DT) based model is used in \cite{liu2018toward} to approximate the Q-values of trained neural network. In that work, the split in DT occurs based on only one feature, and at each terminal node the Q-function is fitted using a  linear model on all features. \cite{verma2018programmatically} uses a \emph{program sketch} $S$ to define the domain of interpretable policies $e$. Interpretable policies are found using a trained black-box oracle $e_N$ as a reference by first conducting a local search in the sketch space $S$ to mimic the behaviour of the oracle $e_N$ and then fine-tuning the policy parameters through online Bayesian optimization. The bias towards generating interpretable programs is done through controlled initialization and local search rather than explicitly capturing \emph{interpretability} as one of the fitness measure.  Particle swarm optimization \cite{kennedy1995particle} is used to generate interpretable fuzzy rule set in \cite{hein2017particle} and is demonstrated on classic control problems involving continuous actions. Works on DT \cite{breiman2017classification} based policies through imitation learning has been carried out in \cite{ross2011reduction}. \cite{bastani2018verifiable} extends this to utilize Q-values and eventually render DT policies involving $< 1,000$ nodes on some toy games and CartPole environment with an ultimate aim to have the induced policies verifiable. \cite{bastani2017interpretability} used axis-aligned DTs to develop interpretable models for black-box classifiers and RL-policies. They first derive a distribution function $\mathcal{P}$ by fitting the training data through axis-aligned Gaussian distributions. $\mathcal{P}$ is then used to compute the loss function for splitting the data in the DT. \cite{vandewiele2016genesim} attempts to generate interpretable DTs from an ensemble using a genetic algorithm. In \cite{ernst2005tree}, regression trees are derived using classical methods such as CART \cite{breiman2017classification} and Kd-tree \cite{bentley1975multidimensional} to model Q-function through supervised training on batch of experiences and comparative study is made with ensemble techniques. In \cite{silva2020optimization}, a gradient based approach is developed to train the DT of pre-fixed topology involving linear split rules. These rules are later simplified to allow only one feature per split node and resulting DTs are pruned to generate simplified rule-set.


While the above methods attempt to generate an interpretable policy, the search process does not use \emph{complexity} of policy in the objective function, instead, they rely on the initializing the search with certain interpretable policies. In our approach described below, we build an efficient search algorithm to directly find relatively interpretable policies as compared to the black-box policies represented using DNN (or tile encoding \cite{sutton1996generalization}) using recent advances in nonlinear optimization. 


%% file: Arxiv_supp_IEEE.tex
\renewcommand{\thefigure}{S-\arabic{figure}}
\renewcommand{\thesection}{S-\roman{section}}
\setcounter{figure}{0}
\renewcommand{\thetable}{A.\arabic{table}}
\setcounter{table}{0}
\setcounter{section}{0}

\section{Data Normalization}
\label{sec:data_normalization}
First, we provide the exact normalization of state variables performed before the open-loop learning task is executed. 
Before training and inducing the non-linear decision tree (NLDT), features in the dataset are normalized using the following equation:
\begin{equation}
    \widehat{x_i} = 1 + (x_i - x_i^{\min})/(x_i^{\max} - x_i^{\min}),
\end{equation}
where $x_i$ is the original value of the $i$-th feature, $\widehat{x_i}$ is the normalized value of the $i$-th feature, $x_i^{\min}$ and $x_i^{\max}$ are minimum and maximum value of $i$-th feature as observed in the training dataset. This normalization will make every feature $x_i$ to lie within $[1,2]$. This is done to ensure that $x_i=0$ is avoided to not cause a division by zero error.

\section{Open Loop NLDT Pruning and Tree Simplification}
The NLDT representing our interpretable AI is induced using successive heirarchical spliting algorithm. A dedicated bilevel approach is used to derive the split rule for each conditional node, i.e., if a child node created after the split is still impure (with its impurity $I > \tau_I$), it is subjected to further split. Initially, we allow the tree to grow to a pre-specified maximum depth of $d_{max}$. The resulting tree is fairly complicated with about hundreds of split nodes. Thus, we simplify this tree further to lower depths and remove redundant splits by pruning them. Lower depth trees are relatively simpler than the full grown depth $d_{max}$ tree and also have better generalizability.

\section{Problems Used in the Study}
In this section, we provide a detail description of the four environments used in this study. 

\subsection{CartPole Environment}
\begin{figure}[hbt]
    \begin{subfigure}{0.46\linewidth}
    \centering\includegraphics[width=\linewidth]{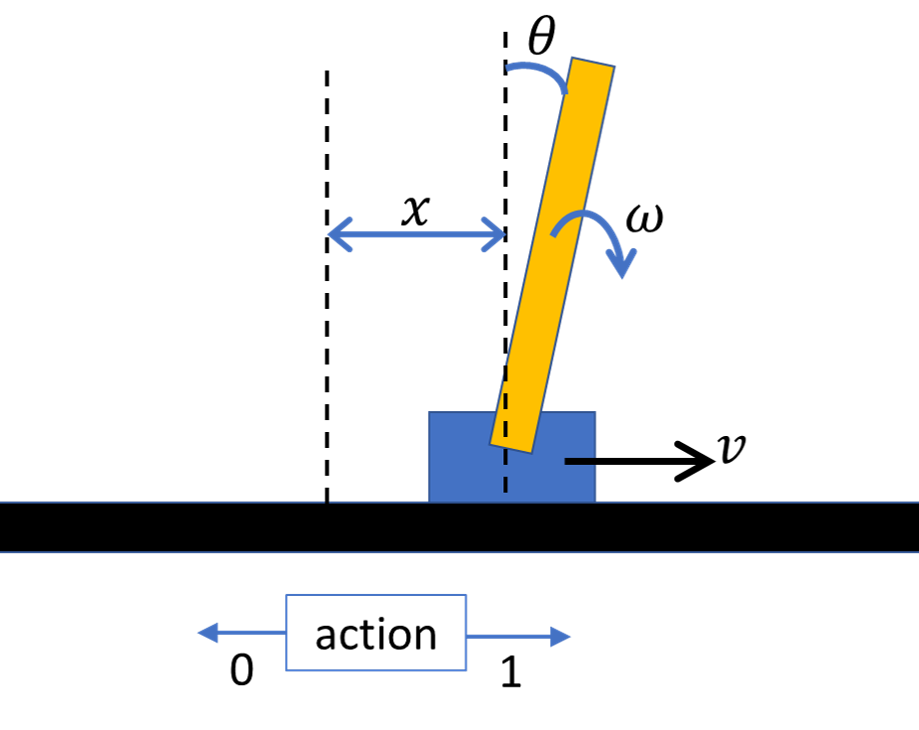}
    \caption{CartPole environment.}
    \label{fig:cart_pole_env}
    \end{subfigure}\ 
    \begin{subfigure}{0.51\linewidth}
    \vspace{-3mm}
    \begin{subfigure}{\linewidth}
    \centering
    \includegraphics[width = \linewidth]{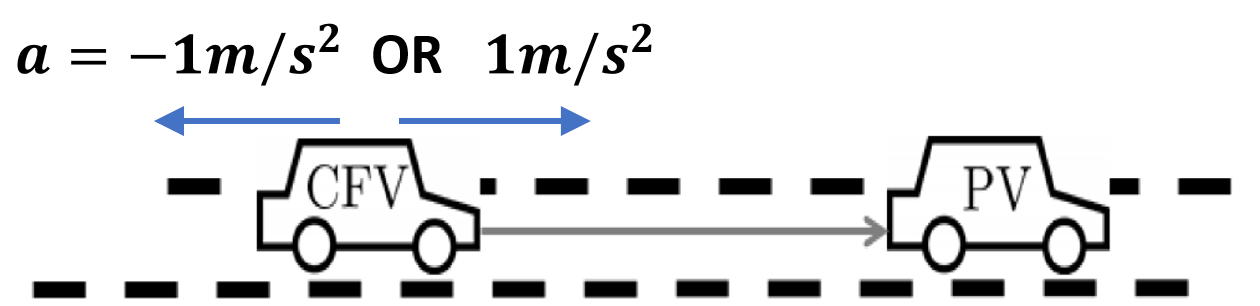}
    \caption{CarFollowing environment.}
    \label{fig:car_following_env}
    \end{subfigure} \\[2mm]
    \begin{subfigure}{\linewidth}
    \centering
    \includegraphics[width = 0.75 \linewidth]{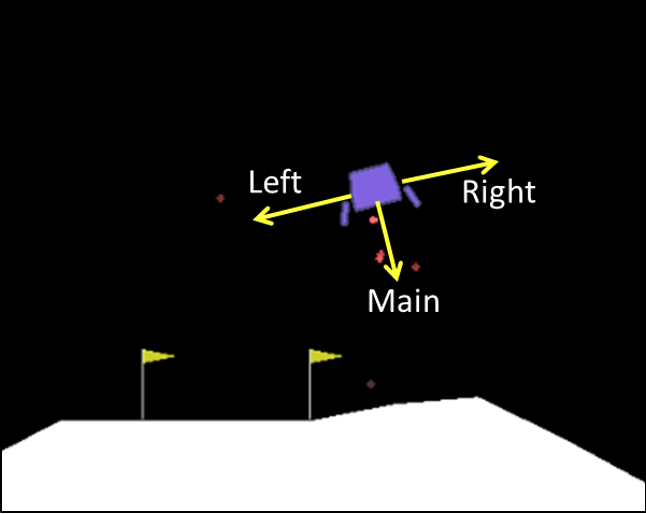}
    \caption{LunarLander environment.}
    \label{fig:lunar_lander_env}
    \end{subfigure}
    \end{subfigure}
    \caption{Other three control problems.}
    \label{fig:3problems}
\end{figure}

The CartPole problem comprises of four state variables: 1) $x$-position ($x \rightarrow x_0$), velocity in +ve $x$ direction ($v \rightarrow x_1$), angular position from vertical ($\theta \rightarrow x_2$) and angular velocity ($\omega \rightarrow x_3$) and is controlled by applying force towards \emph{left} (Action 0) or \emph{right} (Action 1) to the cart (Figure~\ref{fig:cart_pole_env}). The objective is to balance the inverted pendulum (i.e. $-24\deg \le \theta \le 24\deg$) while also ensuring that the cart doesn't fall off from the platform (i.e. $-4.8 \le x \le 4.8$). For every time step, a reward value of 1 is received while $\theta$ is within $\pm 24\deg$. The maximum episode length is set to 200 time steps. 
A deep neural network (DNN) controller is trained on the \emph{CartPole} environment using the PPO algorithm \cite{schulman2017proximal}. 

\subsection{CarFollowing Environment}
\begin{figure}[hbt]
    \centering
    \includegraphics[width = 0.9\linewidth]{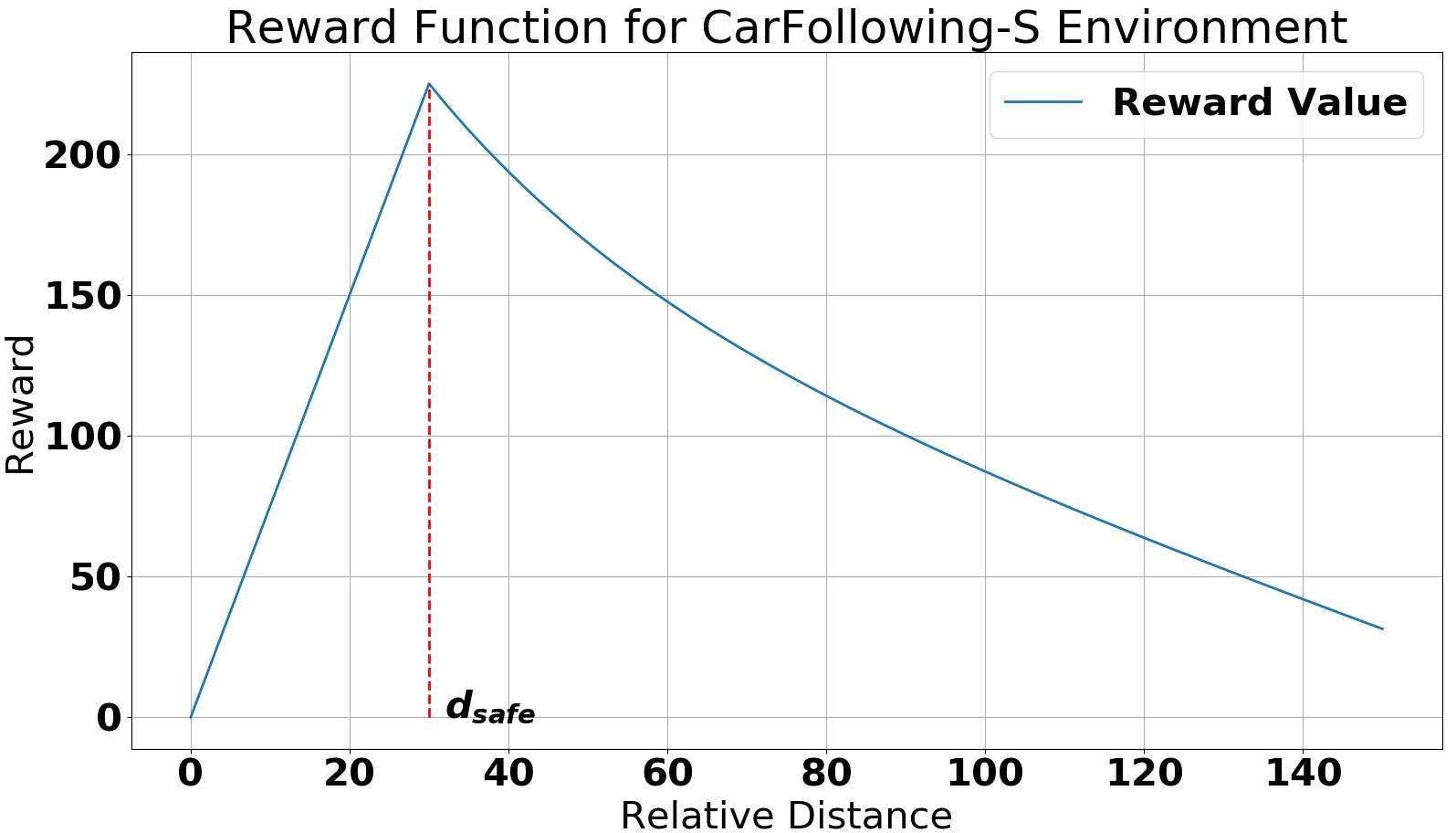}
    \caption{Reward function for CarFollowing environment.}
    \label{fig:car_following_reward}
\end{figure}

As mentioned in the main paper, we have developed a discretized version of the car following problem discussed in \cite{nageshrao2019interpretable} (illustrated in Figure~\ref{fig:car_following_env}), 
wherein the task is to follow the car in the front which moves with a random acceleration profile (between $-1 m/s^2$ and  $+1 m/s^2$) and maintain a safe distance of $d_{safe} = 30 m$  from it. The rear car is controlled using two discrete acceleration values of $+1 m/s^2$ (Action~0) and $-1m/s^2$ (Action~1). The car-chase episode terminates when the relative distance $d_{rel} = x_{front} - x_{rel}$ is either zero (i.e. collision case) or is greater than 150 m. At the start of the simulation, both the cars start with the initial velocity of zero. A DNN policy for CarFollowing problem was obtained using a double Q-learning algorithm \cite{van2015deep}. 
The reward function for the CarFollowing problem is shown in Figure~\ref{fig:car_following_reward}, indicating that a relative distance close to 30 m produces the highest reward. 

It is to note here that unlike the CartPole control problem, where the dynamics of the system was deterministic, the dynamics of the CarFollowing problem is not deterministic due to the random acceleration profile with which the car in the front moves. This randomness introduced by the unpredictable behaviour of the front car makes this problem more challenging.

\subsection{MountainCar Environment}
\begin{figure}[hbtp]
    \centering
    \includegraphics[width=0.6\linewidth]{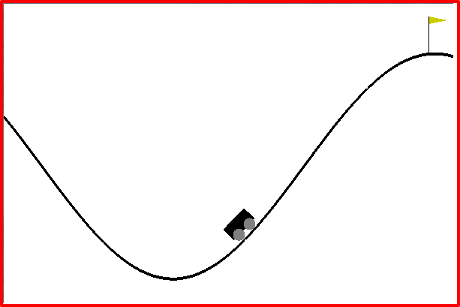}
    \caption{MountainCar Environment.}
    \label{fig:m-car}
\end{figure}

A car starts somewhere near the bottom of the valley and the goal of the task is to reach the flag post located on the right up-hill with non-negative velocity (Figure~\ref{fig:m-car}). The fuel is not enough to directly climb the hill and hence a control strategy needs to be devised to move car back (left up-hill), leverage the potential energy and then accelerate it to eventually reach the flag-post within 200 time steps. The car receives the reward value of $-1$ for each time step, until it reaches the flag-post where the reward value is zero. The car is controlled using three actions: \emph{accelerate left} (Action 0), \emph{do nothing} (Action~1) and \emph{accelerate right} (Action~2) by observing its state which is given by two state-variables: \emph{$x$ position $\rightarrow x_0$} and \emph{velocity $v \rightarrow x_1$}. We use the SARSA algorithm \cite{rummery1994line} with tile encoding to derive the black-box AI controller, which is represented in form of a tensor, which has a total of $151,941$ elements. 

\subsection{LunarLander Environment}
This problem is motivated form a classic problem of design of a rocket-controller. Here, the state of the lunar-lander is expressed with eight state variables, of which six can assume continuous real values, while the rest two are categorical, and can assume a Boolean value (Figure~\ref{fig:lunar_lander_env}). The first six state variables indicate the ($x,y$) position, and velocity and angular orientation and angular velocity of the lunar-lander. The two Boolean state variables provides the indication regarding the left-leg and right-leg contact of lunar-lander with the ground terrain. The lunar-lander is controlled using four actions: Action~0 $\rightarrow$ \emph{do nothing}, Action~1 $\rightarrow$ \emph{fire left engine}, Action~2 $\rightarrow$ \emph{fire main engine} and Action~3 $\rightarrow$ \emph{fire right engine}. The black-box DNN based controller for this problem is trained using the PPO algorithm \cite{schulman2017proximal} and involves two hidden layers of 64 nodes each.

\section{Differences between Open-loop and Closed-loop Searches}
The overall search procedure described in Figure~3 in the main paper clearly indicated that it is a two-step optimization procedure. In the first optimization procedure, an open-loop NLDT (NLDT$_{OL}$) is evolved using a bilevel optimization approach applied recursively to derive split-rule $f(\mathbf{x})$ at each conditional node. Here, each training datapoint consists of a time-instant state-action pair obtained using oracle policy $\pi_{oracle}$. One of the objective function of the overall bilevel algorithm is the minimization of the weighted Gini-score ($F_L$, Eq.~6 in main paper), which quantifies the \emph{purity} of nodes created after the split. This measure can also serve as a \emph{proxy} to indicate the error between predicted action and the AI-model action. For a node $P$, the Gini-score is computed as 
\begin{equation}
    {\rm Gini}(P) = 1 - \mathlarger{\sum_{i = 1}^c \left(\frac{N_i}{N}\right)^2},
    \label{eq:gini}
\end{equation}
where $N$ is the total number of datapoints in node $P$ and $N_i$ is the number of datapoints present in node $P$ which belongs to \emph{action}-$i$. As can be seen from Eq.~\ref{eq:gini}, the computation of Gini-score is computationally cheap and fast. This eventually makes the computation of $F_L$ (Eq.~6 in main paper) to be cheap and fast, a feature which is desired for any bilevel-algorithm since for each solution member in the \emph{upper-level} of the search, a dedicated full run of \emph{lower-level} optimization is performed and if the lower-level objective function is computationally taxing then it will make the overall bilevel algorithm extremely slow. 
Additionally, it is to note here that every rule structure ($f_j(\boldx)$) starting from the root node ($j=0$) is optimized independently by using a subset of the training data dictated by the completed NLDT thus far. Nowhere in the development of the NLDT$_{OL}$, any closed-loop evaluation function (such as, a cumulative reward function of completing the task, or success rate of completion) is used in the optimization process. The structure of the NLDT$_{OL}$ and structure of every rule (with its mathematical structure and coefficients/biases associated with each rule) are evolved. Due to the vastness of the search space of this optimization task, we developed a computationally efficient bilevel optimization procedure composing of a computationally cheap and fast lower-level objective. The two levels allow the structure of each rule and the associated coefficients and biases to be learnt in a hierarchical manner. This is also possible due to recent advances in nonlinear optimization using hybrid evolutionary and point-based local search algorithms \cite{dhebar2020interpretable}. 

On the contrary, the closed-loop optimization restricts its search to a fixed NLDT structure (which is either identical to NLDT$_{OL}$ or a part of it from the root node, as illustrated in Figure~\ref{fig:nldt_lunar_lander_all}), but modifies the coefficients and biases of all rules simultaneously in order to come up with a better closed-loop performance. Here, an entire episode (a series of time-instance state-action pairs from start ($t=0$) to finish ($t=T$)) can be viewed as a \emph{single} datapoint. As an objective function, the average of cumulative-reward collected across 20 episodes, each with a random starting state $S_0$ is used to make a better evaluation of the resulting NLDT. Due to this aspect, the computational burden is more, but the search process stays in a single level. We employ an efficient real-parameter genetic algorithm with standard parameter settings \cite{deb1995simulated,deb2005multi}. To make the search more efficient, we include the NLDT$_{OL}$ (or its part, as the case may be) in the initial population of solutions for the closed-loop search. 

The differences between the two optimization tasks are summarized in Table~\ref{tab:diff}. 
As discussed, both optimization tasks have their role in the overall process. While evaluation of a solution in the open-loop optimization is computationally quicker, it does not use a whole episode in its evaluation process to provide how the resulting rule or NLDT perform on the overall task. The goal here is to maximize the state-action match with the true action as prescribed by $\pi_{oracle}$. This task builds a complete NLDT structure from nothing by finding an optimized rule for every conditional node. The use of a bilevel optimization, therefore, is needed. On the other hand, keeping a part (or whole) of the NLDT$_{OL}$ structure fixed, the closed-loop optimization fine-tunes all associated rules to maximize the cumulative reward $R_{total}$. A closed-loop optimization alone on \emph{episodic} time-instance data to estimate $R_{total}$ will not be computationally tractable in complex problems. 

\begin{table*}[hbt]
\caption{Differences between open-loop and closed-loop optimization problems.}
\label{tab:diff}
\centering\begin{tabular}{|l|p{2.5in}|p{2.5in}|}\hline
Entity & Open-loop Optimization & Closed-loop Optimization \\ \hline
Goal & \begin{tabular}{@{\hspace{1pt}}p{2.5in}}Find each rule-structure $f_j(\boldx)$ one at a time from root node $j=0$\end{tabular} & \begin{tabular}{@{\hspace{1pt}}p{2.5in}}Find overall NLDT simultaneously\end{tabular} \\ \hline
Variables & \begin{tabular}{@{\hspace{1pt}}p{2.5in}}Nonlinear structure $B_{ij}$ for $i$-th term for every $j$-th rule, coefficients $w_{ij}$, and biases $\theta_j$\end{tabular} & \begin{tabular}{@{\hspace{1pt}}p{2.5in}}Coefficients $w_{ij}$ and biases $\theta_j$ for all rules ($j$) in the NLDT\end{tabular} \\ \hline
Each training data & \begin{tabular}{@{\hspace{1pt}}p{2.5in}}State-action pair ($\boldx^t$-$\bolda^t$) for each time-instance $t$\end{tabular} & \begin{tabular}{@{\hspace{1pt}}p{2.5in}}Randomly initialized $M$ Episodes comprising of state-action-reward triplets ($\boldx^t$-$\bolda^t$-$r_t$, for $t=1,\ldots,T$) for each simulation\end{tabular} \\ \hline
Objective function & \begin{tabular}{@{\hspace{1pt}}p{2.5in}}Weighted Gini-score (mismatch in actions)\end{tabular} &  \begin{tabular}{@{\hspace{1pt}}p{2.5in}}Average cumulative reward value\end{tabular} \\ \hline
Optimization method & \begin{tabular}{@{\hspace{1pt}}p{2.5in}}Bilevel optimization: Upper-level by customized evolutionary algorithm and lower-level by regression\end{tabular} & \begin{tabular}{@{\hspace{1pt}}p{2.5in}}Single-level genetic algorithm\end{tabular} \\ \hline
Termination condition & \begin{tabular}{@{\hspace{1pt}}p{2.5in}}
     Upper level (Change in fitness $< 0.01\%$ for consecutive 5 generations in Upper level GA, with maximum 100 generations).\\
     Lower level (Change in fitness $< 0.01\%$ for consecutive 5 generations, with maximum 50 generations).
\end{tabular}   & 30 generations \\ \hline
Outcome & NLDT$_{OL}$ & NLDT* \\ \hline
\end{tabular}
\end{table*}

\section{Additional Results}

\subsubsection{NLDT for CarFollowing Problem}
\begin{figure}[hbt]
    \centering
    \includegraphics[width = 0.8\linewidth]{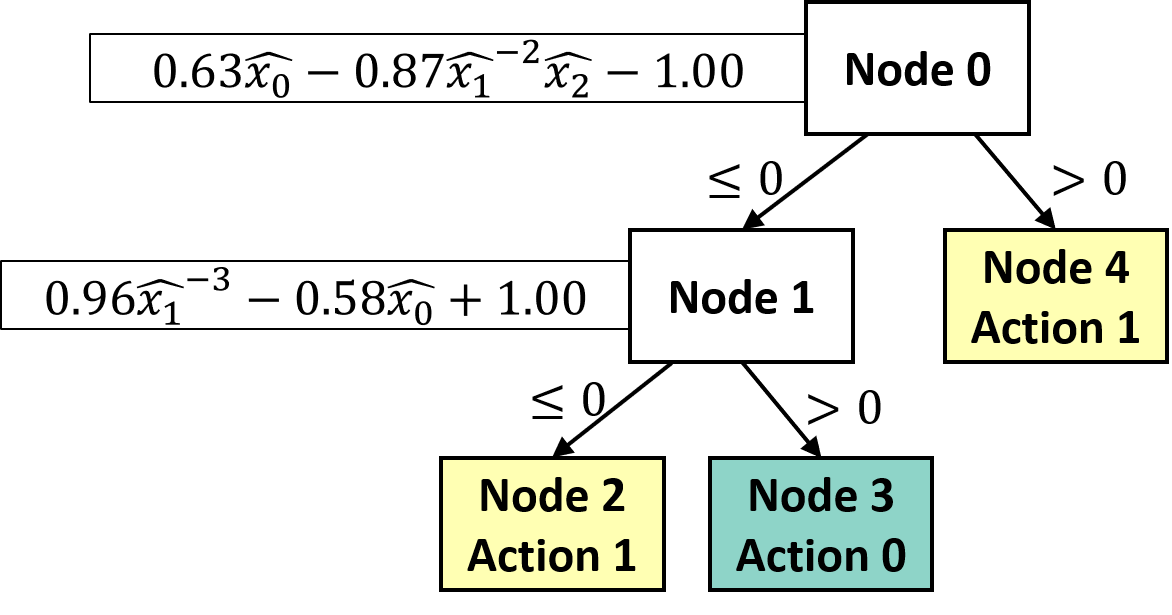}
    \caption{NLDT$_{OL}$ for the CarFollowing problem. Normalization constants are: $x^{\min}$ = [0.25, -7.93, -1.00],
 $x^{\max}$ = [30.30, 0.70, 1.00].}
    \label{fig:nldt_CarFollowing}
\end{figure}

The NLDT$_{OL}$ obtained for the CarFollowing problem is shown in Figure~\ref{fig:nldt_CarFollowing}. 
The rule-set is provided in its natural if-then-else form in Algorithm~\ref{algo:CarFollowing}.

\begin{algorithm}
\addtocontents{loa}{\vskip 14.4pt}
\caption{Ruleset corresponding to NLDT$_{OL}$ (Figure~\ref{fig:nldt_CarFollowing}) of the CarFollowing problem.}
\label{algo:CarFollowing}
\eIf{$0.63 \widehat{x_{0}} - 0.87 \widehat{x_{1}}^{-2}\widehat{x_{2}} - 1.00 \le 0$}
{
\eIf{$0.96 \widehat{x_{1}}^{-3} - 0.58 \widehat{x_{0}} + 1.00 \le 0$}{Action = 1}{Action = 0}
}
{Action = 1}
\end{algorithm}

Recall that the physical meaning of state variables is: $x_0 \rightarrow d_{rel}$ (relative distance between front car and rear car), $x_1 \rightarrow v_{rel}$ (relative velocity between front car and rear car) and $x_2 \rightarrow a$ (acceleration value ($-1$ or $+1$ m/s$^2$) at the previous time step). Action = 1 stands for acceleration and Action = 0 denotes deceleration of the rear car in the next time step. 

From the first rule (Node~0), it is clear that if the rear car is close to the front car ($\widehat{x_0} \approx 1$), the root function $f_0(\boldx)$ is never going to be positive for any value of relative velocity or previous acceleration of the rear car (both $\widehat{x_1}$ and $\widehat{x_2}$ lying in [1,2]). Thus, Node~4 (Action = 1, indicating acceleration of the rear car in the next time step) will never be invoked when the rear car is too close to the front car. Thus for $\widehat{x_0} \approx 1$, the control always passes to Node~1. A little analysis will also reveal that for $\widehat{x_0} \approx 1$, the rule $f_1(\boldx) > 0$ for any relative velocity $\widehat{x_1} \in [1,2]$. This means that when the two cars are relatively close, only Node~3 gets fired to decelerate (Action = 0) the rear car. This policy is intuitively correct, as the only way to increase the gap between the cars is for the controlled rear car to be decelerating. 

However, when the rear car is far away for which $\widehat{x_0} \approx 2$, Action~1 (Node~4) gets fired if $\widehat{x_1} > 1.829\sqrt{\widehat{x_2}}$. If the rear car was decelerating in the previous time step (meaning $\widehat{x_2} = 1$), the obtained NLDT recommends that the rear car should accelerate if $\widehat{x_1} \in [1.829, 2]$, or when the magnitude of the relative velocity is small, or when $x_1 \in [-0.776, 0.700] m/s$. This will help maintain the requisite distance between the cars. On the other hand, if the rear car was already accelerating in the previous time step ($\widehat{x_2} = 2$), Node~4 does not fire, as $\widehat{x_1}$ can never be more than $1.829\sqrt{2}$ and the control goes to Node~1 for another check. Thus, the rule in Node~0 makes a fine balance of the rear car's movement to keep it a safe distance away from the front car, based on the relative velocity, position, and previous acceleration status. When the control comes to Node~1, Action~1 (acceleration) is invoked if $\widehat{x_1} \geq 0.96/(0.58\widehat{x_0}-1)$. For $\widehat{x_0} \approx 2$, this happens when $\widehat{x_1} > 1.817$ (meaning that when the magnitude of the relative velocity is small, or $x_1 \in [-0.879, 0.700] m/s$), the rear car should accelerate in the next time step. For all other negative but large relative velocities $x_1 \in [-7.930, 0.879] m/s$), meaning the rear car is rushing to catch up the front car, the rear car should decelerate in the next time step. From the black-box AI data, our proposed methodology is able to obtain a simple decision tree with two nonlinear rules to make a precise balance of movement of the rear car and also allowing us to understand the behavior of a balanced control strategy. 

\begin{figure*}[htbp]
    \centering
    \includegraphics[width=0.95\linewidth]{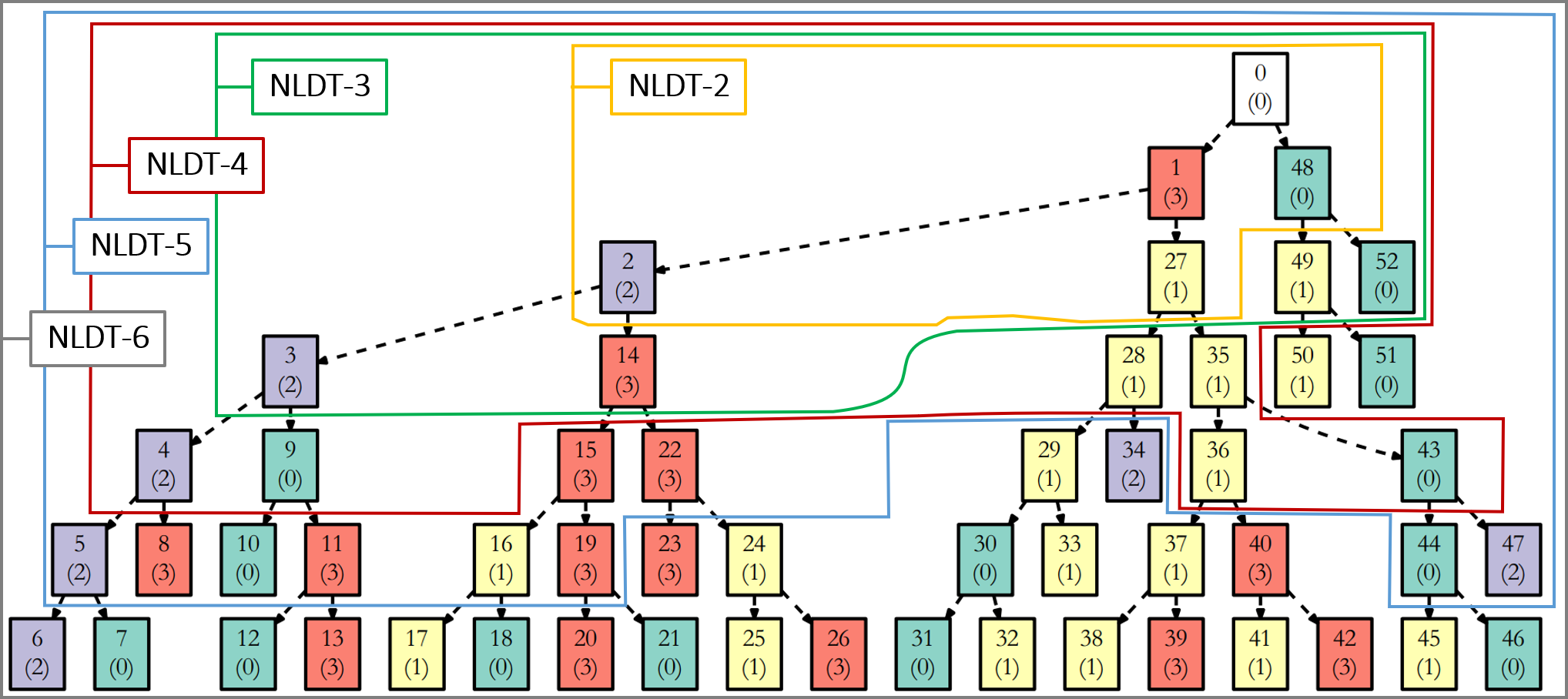}
    \caption{NLDT-6 (with 26 rules) and other lower depth NLDTs for the LunarLander problem. Lower depth NLDTs are extracted from the depth-6 NLDT. Each node has an associated node-id (on top) and a node-class (mentioned in bottom within parenthesis). Table~6 in main paper provides results on closed-loop performance obtained using these trees \emph{before} and \emph{after} applying re-optimization on rule-sets using the closed-loop training procedure.}
    \label{fig:nldt_lunar_lander_all}
\end{figure*}

\subsection{LunarLander Problem}
One of the NLDT$_{OL}$s induced using the open-loop supervised training is shown in Figure~\ref{fig:nldt_lunar_lander_all}. The performance of this NLDT$_{OL}$ was presented in the main paper. It has a depth of 6 and it involves a total of 26 rules. 
The figure also shows how this 26-rule NLDT$_{OL}$ can be pruned to smaller sized NLDTs (such as, NLDT-5, NLDT-4, NLDT-3, NLDT-2) starting from the root node. A compilation of results corresponding to these trees regarding their closed-loop performance before and after re-optimizing them using the closed-loop training is shown in Table~6 of the main paper. The main paper has also presented a four-rule NLDT*-3 obtained by a closed-loop training of the above NLDT-3.

\begin{figure}[htbp]
    \centering
    \includegraphics[width = 0.8\linewidth]{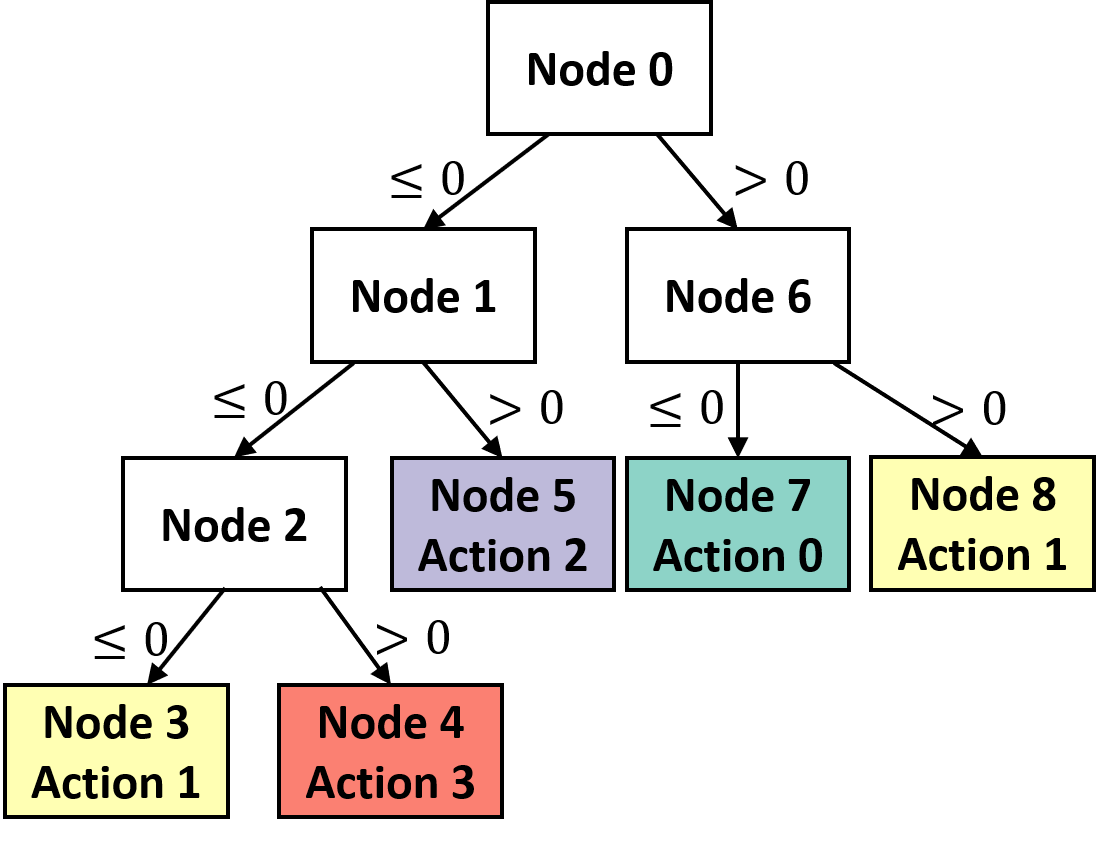}
    \caption{Topology of Depth-3 NLDT$^{(P)}_{OL}$ obtained from a different run on the LunarLander problem. The equations corresponding the conditional-nodes before and after re-optimization are provided in Table~\ref{tab:LunarLander_new_rules}.}
    \label{fig:nldt_LunarLander_new}
\end{figure}

To demonstrate the efficacy and repeatability of our proposed approach, we perform another run of the open-loop and closed-loop training and obtain a slightly different NLDT*-3, which is shown in Figure~\ref{fig:nldt_LunarLander_new}.
This NLDT also has four rules, which are shown in  Table~\ref{tab:LunarLander_new_rules}. Four rules rules at the pruned NLDT$^{(P)}_{OL}$ (Depth 3) are also shown in the table for a comparison.
\begin{table*}[htbp]
\setcellgapes{5pt}
\makegapedcells
\centering
\caption{NLDT rules before and after the closed-loop training for LunarLander problem, for which NLDT* is shown in Figure~\ref{fig:nldt_LunarLander_new}. Video at {\textcolor{blue}{\url{https://youtu.be/DByYWTQ6X3E}}} shows the simulation output of the performance of NLDTs with rule-sets mentioned in this table. Respective minimum and maximum state variables are
$x^{\min}$ = [-0.38, -0.08, -0.80, -0.88, -0.42, -0.85, 0.00, 0.00],
 $x^{\max}$ = [0.46, 1.52, 0.80, 0.50, 0.43, 0.95, 1.00, 1.00], respectively.}
 \label{tab:LunarLander_new_rules}
\begin{tabular}{|c|c|}\hline
{\bf Node} & {\bf Rules before Re-optimization (Depth-3 NLDT$^{(P)}_{OL}$)}\\\hline
0 & $\left| - 0.23 \widehat{x_{0}}\widehat{x_{2}}^{-1}\widehat{x_{6}}^{-1}\widehat{x_{7}}^{-1} - 1.00 \widehat{x_{1}}^{-1}\widehat{x_{6}} - 0.79 \widehat{x_{0}}^{-1}\widehat{x_{1}}^{-1}\widehat{x_{6}}^{2} + 0.83 \right| - 0.85 $ \\\hline 
1 & $0.17 \widehat{x_{2}}^{-1} - 0.64 \widehat{x_{3}}\widehat{x_{7}}^{-1} + 0.90 \widehat{x_{1}}^{-2}\widehat{x_{6}}^{-2}\widehat{x_{7}}^{-3} + 0.29 $ \\\hline 
2 & $0.82 \widehat{x_{7}}^{-1} + 0.52 \widehat{x_{0}}^{-1}\widehat{x_{4}}\widehat{x_{6}}^{-1} - 0.59 \widehat{x_{4}}^{-1} - 0.95 $ \\\hline 
6 & $\left| - 0.16 \widehat{x_{4}}^{-3}\widehat{x_{6}}^{-3}\widehat{x_{7}} - 0.86 \widehat{x_{0}}\widehat{x_{5}}^{-1}\widehat{x_{6}}^{-3} + 1.00 \widehat{x_{4}}\widehat{x_{6}}^{-1} - 0.70 \right| - 0.26 $ \\\hline 
\end{tabular}
\vspace{5pt}
\begin{tabular}{|c|c|}\hline
{\bf Node} & {\bf Rules after Re-optimization (Depth-3 NLDT*)}\\\hline
0 & $\left| - 0.39 \widehat{x_{0}}\widehat{x_{2}}^{-1}\widehat{x_{6}}^{-1}\widehat{x_{7}}^{-1} - 0.96 \widehat{x_{1}}^{-1}\widehat{x_{6}} - 0.12 \widehat{x_{0}}^{-1}\widehat{x_{1}}^{-1}\widehat{x_{6}}^{2} + 0.89 \right| - 0.80 $ \\\hline 
1 & $0.17 \widehat{x_{2}}^{-1} - 0.78 \widehat{x_{3}}\widehat{x_{7}}^{-1} + 0.90 \widehat{x_{1}}^{-2}\widehat{x_{6}}^{-2}\widehat{x_{7}}^{-3} + 0.35 $ \\\hline 
2 & $0.82 \widehat{x_{7}}^{-1} + 0.52 \widehat{x_{0}}^{-1}\widehat{x_{4}}\widehat{x_{6}}^{-1} - 0.59 \widehat{x_{4}}^{-1} - 0.96 $ \\\hline 
6 & $\left| - \left(1.3 \times 10^{-3}\right)\widehat{x_{4}}^{-3}\widehat{x_{6}}^{-3}\widehat{x_{7}} - 0.86 \widehat{x_{0}}\widehat{x_{5}}^{-1}\widehat{x_{6}}^{-3} + 0.65 \widehat{x_{4}}\widehat{x_{6}}^{-1} - 0.42 \right| - 0.26 $ \\\hline 
\end{tabular}
\end{table*}

It can be noticed that the re-optimization of NLDT through closed-loop training (Section~5.2 in main paper)) modifies the values of coefficients and biases, however the basic structure of all four rules remains intact. 

\subsection{10-link ManAction Vs Time Plots}
\begin{figure*}[t!]
    \centering
    \begin{subfigure}[b]{0.3\textwidth}
         \centering
         \includegraphics[width=\textwidth]{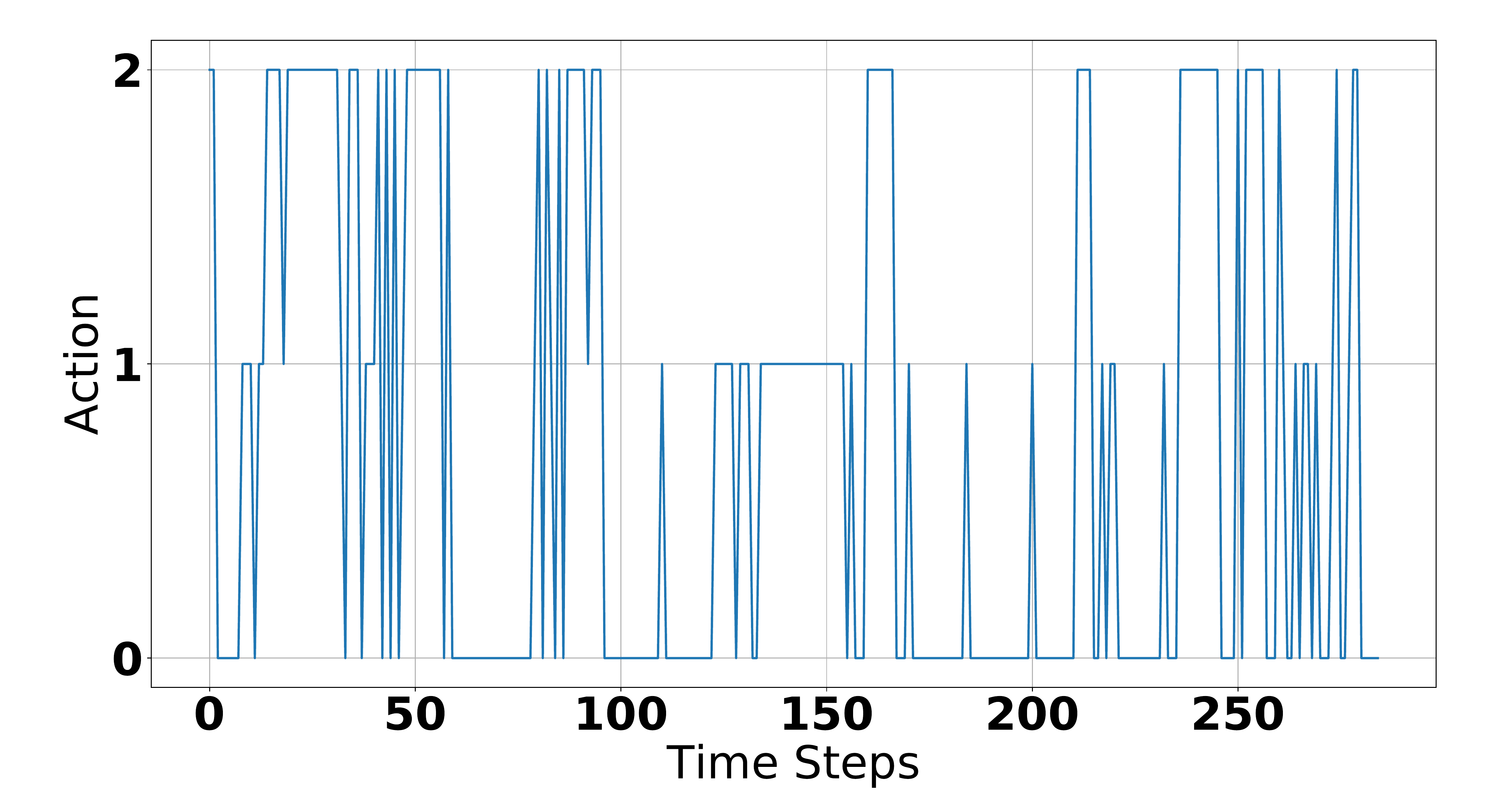}
         \caption{DNN}
         \label{fig:10_link_DNN}
     \end{subfigure}
     \hfill
     \begin{subfigure}[b]{0.3\textwidth}
         \centering
         \includegraphics[width=\textwidth]{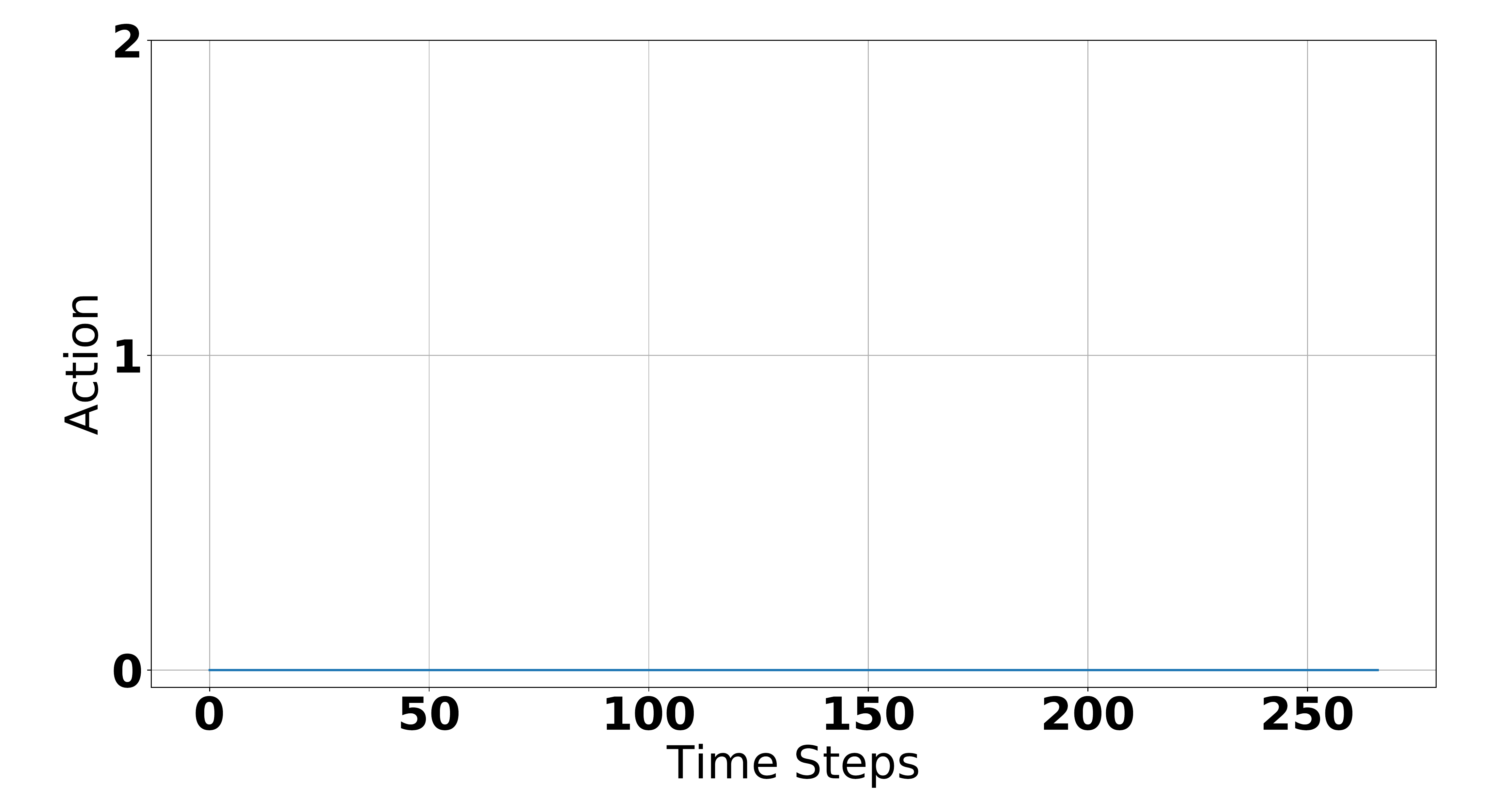}
         \caption{NLDT* (SQP)}
         \label{fig:10_link_SQP}
     \end{subfigure}
     \hfill
     \begin{subfigure}[b]{0.3\textwidth}
         \centering
         \includegraphics[width=\textwidth]{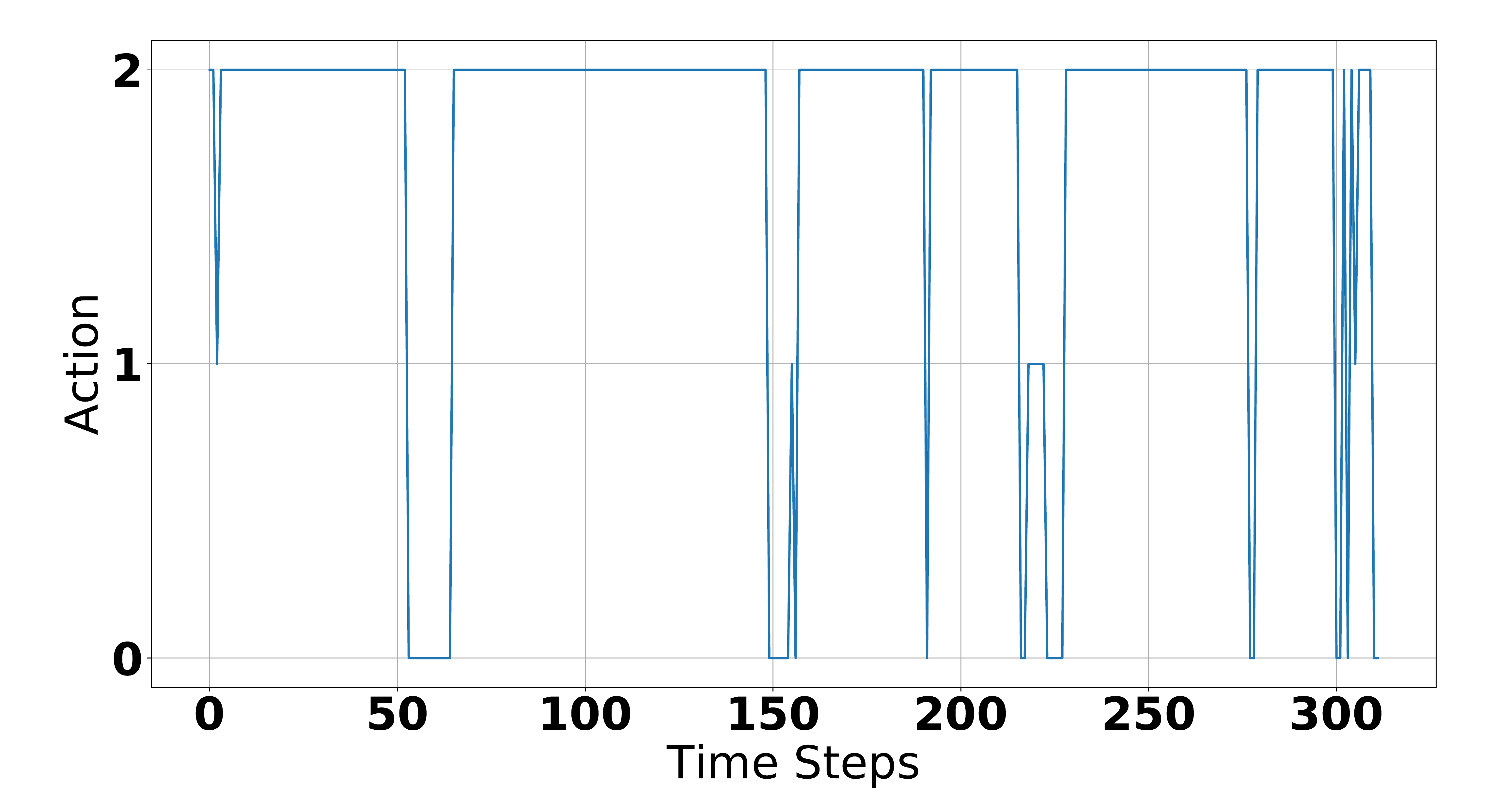}
         \caption{NLDT* (RGA)}
         \label{fig:10_link_RGA}
     \end{subfigure}
    \caption{Action Vs. Time plot for 10-Link manipulator problem. Figure~\ref{fig:10_link_SQP} provides the plot for NLDT* which is obtained from the NLDT$_{OL}$ trained using SQP algorithm in lower-level. Similarly, Figure~\ref{fig:10_link_RGA} provides the plot for NLDT* which is obtained from the NLDT$_{OL}$ trained using RGA algorithm in lower-level.}
    \label{fig:10_link_plots}
\end{figure*}

Action Vs. time plot obtained using NLDT* (SQP), NLDT* (RGA) and DNN for a 10-link serial manipulator problem is shown in Figure~\ref{fig:10_link_plots}

This is a slightly difficult problem to solve than the 5-link version since it involves twice the number of state variables. Interestingly, the search for NLDT* provides us with the simplest solution to this problem to lift the end-effector to the desired height of 2 units above the base. The simplest solution here is to give a constant torque in one direction as shown in Figure~\ref{fig:10_link_SQP}. However, for this problem the best closed-loop performance in terms of cumulative reward (see Table~8 in main paper) is obtained using the control strategy corresponding to NLDT* (RGA) (Figure~\ref{fig:10_link_RGA}). Here too, for most of the states, only one action is required, and occasionally other actions are invoked.